\documentclass[journal,12pt,onecolumn,draftclsnofoot]{IEEEtran}

\usepackage{cite}
\usepackage{graphicx}
\graphicspath{{./Images/}}

\usepackage[cmex10]{amsmath}

\usepackage{float}
\usepackage{verbatim}
\usepackage{relsize,bbm}
\usepackage{amsfonts}
\usepackage{amsthm}
\usepackage{amssymb}
\usepackage{latexsym}
\usepackage{url}
\usepackage{epstopdf}
\usepackage{enumerate}
\usepackage{mathrsfs}
\usepackage{tabulary, hyperref}
\usepackage{color}
\usepackage{balance}
\usepackage{amsmath}
\usepackage{cases}

\DeclareMathOperator*{\argmax}{arg\,max}
\DeclareMathOperator*{\argmin}{arg\,min}
\newtheorem{theorem}{Theorem}
\newtheorem{lemma}{Lemma}

\newtheorem{proposition}{Proposition}

\newcommand{\bbP}{{\mathbb P}}
\newcommand{\bbR}{{\mathbb R}}
\newcommand{\bbE}{{\mathbb E}}

\newcommand{\one}{\mathbbm{1}}

\newcommand{\cA}{{\cal A}}

\newcommand{\cB}{{\cal B}}
\newcommand{\cP}{{\cal P}}
\newcommand{\cS}{{\cal S}}
\newcommand{\cX}{{\cal X}}
\newcommand{\cY}{{\cal Y}}

\newcommand{\cU}{{\cal U}}
\newcommand{\cD}{{\cal D}}
\newcommand{\cG}{{\cal G}}
\newcommand{\cV}{{\cal V}}
\newcommand{\cE}{{\cal E}}
\newcommand{\cT}{{\cal T}}

\newcommand{\dP}{\zeta_P}

\newcommand{\ML}{{\mathrm{ML}}}
\newcommand{\brx}{{\boldsymbol{\mathrm{x}}}}
\newcommand{\bry}{{\boldsymbol{\mathrm{y}}}}

\newcommand{\nbd}{{\mathrm{nbd}}}
\newcommand{\pa}{{\mathrm{pa}}}
\newcommand{\CL}{{\mathrm{CL}}}
\newcommand{\BK}{{\mathrm{BK}}}
\newcommand{\NKS}{{\mathrm{NKS}}}

\newcommand*{\medcup}{\mathbin{\scalebox{1.5}{\ensuremath{\cup}}}}%

\allowdisplaybreaks

\begin{document}
		
\title{Exact Asymptotics for Learning Tree-Structured Graphical Models with Side Information: Noiseless and Noisy Samples}

\author{Anshoo Tandon,~\IEEEmembership{Member,~IEEE}, Vincent Y.\ F.\ Tan,~\IEEEmembership{Senior Member,~IEEE}, \\ and Shiyao Zhu
	\thanks{This work was supported in part by a Singapore Ministry of Education Tier 2 grant (R-263-000-C83-112).}%
	\thanks{A.~Tandon and S.~Zhu are with the Department of Electrical and Computer Engineering, National University of Singapore, Singapore 117583 (email: anshoo.tandon@gmail.com, shiyao.sg@gmail.com).}%
	\thanks{V.~Y.~F.~Tan is with the Department of Electrical and Computer Engineering, and with the Department of Mathematics, National University of Singapore, Singapore (email: vtan@nus.edu.sg).}%
}

\maketitle

\begin{abstract}
Given side information that an Ising tree-structured graphical model is homogeneous and has no external field, we derive the exact asymptotics of learning its structure from independently drawn samples. 
Our results, which leverage the use of probabilistic tools from the theory of strong large deviations, refine the large deviation (error exponents) results of Tan, Anandkumar, Tong, and Willsky [IEEE Trans.\ on Inform.\ Th., 57(3):1714--1735, 2011] and strictly improve those of Bresler and Karzand [Ann.\ Statist., 2020].  In addition, we extend our results to the scenario in which the samples are observed in random noise. In this case, we show that they strictly improve on the  recent results of Nikolakakis, Kalogerias, and Sarwate [Proc.\ AISTATS, 1771--1782, 2019]. Our theoretical results demonstrate keen agreement with   experimental results for sample sizes as small as that in the hundreds. 
\end{abstract}


\section{Introduction}
The learning of graphical models~\cite{Wainwright08_FnT} from data samples is an important and fundamental task in statistical inference and learning. Graphical models provide a robust framework for capturing the statistical dependencies among a large collection of random variables and derive their power from their ability to provide a diagrammatic representation of a multivariate distribution in the form of a graph. The edges of the graphical model encode conditional independence relations amongst a set of random variables. Graphical models have found extensive applications in image denoising~\cite{Besag86},
iterative decoding~\cite{Kschischang98}, natural language processing~\cite{Navigli10_PAMI}, and optimization~\cite{Maneva07}. See~\cite{Wainwright08_FnT} for a comprehensive exposition of learning and inference in graphical models. 

The task of learning graphical models entails using a set of independently drawn samples to infer the underlying set of edges of the graph. The most basic algorithm is the Chow-Liu algorithm~\cite{ChowLiu68} which finds the tree-structured graphical model that is closest in the Kullback-Leibler (KL) divergence sense (more precisely, the reverse I-projection~\cite{Csiszar03_IP} sense) to the empirical distribution of the samples. Equivalently, the Chow-Liu algorithm is a maximum likelihood (ML) rule---one that maximizes the likelihood of the observed samples over all candidate tree models. If the samples are known to be generated from a particular tree-structured model, then an early result by Chow and
Wagner~\cite{ChowWagner73} shows that the Chow-Liu algorithm is consistent in the sense that as the number of samples tends to infinity, the set of edges of the tree is recovered with overwhelming probability. Using large deviations theory~\cite{HollanderBook00}, and in particular Sanov's theorem~\cite[Ch.~11]{CoverBook06},\, Tan, Anandkumar, Tong, and Willsky~\cite{Tan11IT} quantified the decay rate of the error probability in terms of a quantity known as the {\em error exponent}.

While the error exponent is a useful quantification of the ease or difficulty of learning tree-structured graphical models~\cite{Tan10TSP}, empirical evidence provided in~\cite{Tan11IT} shows that the estimates of the error probability when the number of samples is small is poor. Motivated
by practical scenarios in which the number of samples is relatively small, in this paper, we adopt a new framework for estimating the probability of error in learning trees. We adopt the probabilistic theory of strong large deviations~\cite{BahadurRao60,BlackwellHodges59} to obtain the {\em exact asymptotics} of learning a certain simple class of Ising tree models---namely, 
that with no external field and are homogeneous. Using these two pieces of side information, we develop a new and optimal rule for learning tree models. We show that the approximations of the error probability are not only easily computable, they are also extremely accurate at small samples sizes (of the order of hundreds of samples). Thus, our framework serves as a powerful and useful refinement of the results of~\cite{Tan11IT}  and more recent work by Bresler and Karzand~\cite{BreslerKarzand18}. In fact, we show that the exponent we obtain is at least $3$ times larger than that of~\cite[Sec.~7.2]{BreslerKarzand18}. An auxiliary contribution here is the quantification
of the improvement of the tie-breaking rule in the implementation of the maximum weight spanning tree (MWST) procedure to learn the tree given edge weights over a na\"{i}ve, conservative, and pessimistic rule in which a decoding error is immediately declared whenever two pairs of
nodes have the same weight. This is analogous to the quantification of the advantage of tie-breaking in the random coding union (RCU) bound for channel coding~\cite{Haim18}.

En route to using the theory of strong large deviations to obtain estimates of the error probability for learning the above mentioned class of Ising models, we also find that our newly-developed analytical tools are also useful in estimating the error probability of learning tree models with {\em noisy} samples, a setting recently studied in a series of works by Nikolakakis, Kalogerias, and Sarwate~\cite{Nikolakakis19AISTATS,Nikolakakis19Predictive,Nikolakakis19NonParametric}. This problem setup in which we observe noisy samples is particularly pertinent in scenarios in which measurement errors are introduced in systems with limited precision, e.g., a sensor network with faulty receivers or a biological
system with diagnostic errors in differentiating malignant and benign biological cells. We show that for the class of Ising models under consideration, the error exponent of our learning rule is {\em optimal}. Thus, our decoding rule and accompanying analysis result in significantly
improved error probability and sample complexity estimates compared to that by Nikolakakis, Kalogerias, and Sarwate~\cite{Nikolakakis19AISTATS,Nikolakakis19Predictive,Nikolakakis19NonParametric}. Again, we show through   numerical experiments that our easily computable approximations of the error probability are in keen agreement with the empirical observations. We note that this setting is in contrast to recent work on robust tree learning in \emph{adversarial} noise~\cite{Cheng18NeurIPS}; in our work, \emph{random} noise is added to clean samples.

 The rest of this paper is structured as follows: In Sec.~\ref{sec:prelim}, we describe some preliminaries on graphical models and state the problem precisely. In Sec.~\ref{sec:ee}, we review the ML procedure for learning tree models with and without side information~\cite{ChowLiu68}, and the error exponent analysis in~\cite{Tan11IT}. In Sec.~\ref{sec:ExactAsymptotics}, we present our main contribution---an exact asymptotic result  for learning trees with side information. In Sec.~\ref{sec:ExactAsymptotics_NS}, we leverage the preceding analysis to study the performance of tree learning when the samples are observed in noise. Simulation results to corroborate the theory are presented in Sec.~\ref{sec:NumericalResults}. Finally, we wrap up the discussion and present various avenues for further research in Sec.~\ref{sec:reflect}.

\section{Preliminaries and Problem Statement} \label{sec:prelim}
An {\em undirected graphical model}, also known as a {\em Markov random field}, is a multivariate  probability distribution that factorizes according to the structure of the given undirected graph~\cite{LauritzenBook}. Specifically, a $p$-dimensional random vector $\brx := [x_1, \ldots, x_p]$ is said to be \emph{Markov} on an undirected graph $\cG = (\cV, \cE)$ with vertex (or node) set $\cV = \{1,\ldots,p\}$ and edge set $\cE \subset \binom{\cV}{2}$ if its distribution $P(\brx)$ satisfies the (local) Markov property $P(x_i | x_{\cV \setminus i}) = P(x_i | x_{\nbd(i)})$ where $\nbd(i) := \{j \in \cV : \{i,j\} \in \cE\}$ is the {\em neighborhood} of node $i$. 

This paper focuses on tree-structured graphical models $P$, where the underlying graph of $P$ is an acyclic and connected graph, denoted by $T_P = (\cV, \cE_{P})$ with $|\cE| = p-1$. Tree-structured graphical models factorize as~\cite{LauritzenBook}
\begin{equation*}
P(\brx) = \prod_{i \in \cV} P_i(x_i) \prod_{\{i,j\} \in \cE_{P}} \frac{P_{i,j}(x_i,x_j)}{P_i(x_i) P_j(x_j)} ,
\end{equation*} 
where $P_i$ and $P_{i,j}$ are the marginals on node $i \in \cV$ and edge $\{i,j\} \in \cE_{P}$, respectively. For an undirected tree, we may assume, without loss of generality, that node $1$ is the \emph{root} node, and arrange all the nodes at different levels on a plane, with node $1$ at level-$0$. Then, the graphical model $P$ can be alternatively factored as~\cite{ChowLiu68}
\begin{equation}
P(\brx) = P_1(x_1) \prod_{i=2}^p P_{i|\pa(i)}(x_i | x_{\pa(i)}) , \label{eq:TreeFactorizationP}
\end{equation} 
where $\pa(i)$ denotes the unique parent node of node $i$ and $P_{i|\pa(i)}$ is the conditional distribution of node (or variable) $i$ given node $\pa(i)$.

\subsection{System Model} \label{sec:SystemModel}
In our study, we consider binary random variables with alphabet $\cX = \{0,1\}$. We further assume that the tree-structured graphical model $P$, for $p > 2$ nodes, has the following properties:
\begin{enumerate} 
	\item[P1] (\emph{Zero external field}): The marginals are uniform, i.e., $P_i(0) = P_i(1) = 0.5$, for $1 \le i \le p$.
	\item[P2] (\emph{Homogeneity}): For every edge $\{i,j\} \in \cE_{P}$, we have $P_{i,j}(0,1) = P_{i,j}(1,0) = \theta/2$, where $\theta$ lies in the open interval $(0,0.5)$.
\end{enumerate}
Since the multiplicative group $\{+1,-1\}$ is isomorphic to the additive group $\{0,1\}$~\cite{HersteinBook75}, it is seen that property P1 corresponds to \emph{Ising models} with zero external field, and is a common assumption in related literature on learning tree-structured graphical models~\cite{BreslerKarzand18, Nikolakakis19AISTATS, Nikolakakis19Predictive}. Properties P1 and P2 help to make the analysis tractable and serve to capture the essential features of a simplified tree model. In a related work, lower bounds on the sample complexity for learning Ising models, satisfying properties P1 and P2, were presented in~\cite{Shanmugam14NeurIPS}. Multiple applications of the Ising model to real-world problems, ranging from understanding the statistical mechanics of social dynamics to modeling price trends in financial markets, were discussed in~\cite{Nikolakakis19AISTATS}. The problem of high-dimensional Ising model selection was analyzed in~\cite{Anandkumar12}, while also discussing a concrete example of the application of the Ising model to U.S. senate voting network.

Let $\cT^p$ denote the set of all distinct trees with $p$ nodes, and let $\cD(\cT^p, \theta)$ denote the set of all tree distributions $P$ satisfying properties P1 and P2. Note that property P2 implies a positive correlation between nodes connected by an edge. Therefore, if $\{i,j\} \in \cE_{P}$, then $x_i$ and $x_j$ are more likely to be similar (rather than dissimilar), and we have $P_{i,j}(0,0) + P_{i,j}(1,1) = 1-\theta > \theta = P_{i,j}(0,1) + P_{i,j}(1,0)$. In the following, we will denote the set of all probability distributions over the alphabet $\cX^p$ by $\cP(\cX^p)$, and we take the natural base for logarithms.

\subsection{Problem Statement}
We consider the problem of tree learning with side information where
we are given $n$ i.i.d. $p$-dimensional samples $\brx^n := \{\brx_1,\ldots,\brx_n\}$ from an unknown tree-structured graphical model $P$ satisfying P1 and P2. The side information to the tree learning algorithm is the knowledge that $P$ satisfies P1 and P2. For $1 \le k \le n$, each sample or observation $\brx_k := [x_{k,1},\ldots,x_{k,p}]$  is a vector of $p$ dimensions, where $x_{k,j} \in \cX = \{0,1\}$ for $1 \le j \le p$.

Given $\brx^n$, the ML  estimator of the unknown distribution $P$ is
\begin{equation}
P_{\ML}(\brx^n) := \argmax_{Q \in \cD(\cT^p, \theta)} \sum_{k=1}^n \log Q(\brx_k) . \label{eq:P_ML_v1}
\end{equation}
We denote the tree graph of the ML estimate $P_{\ML}(\brx^n)$ by $T_{\ML}(\brx^n) = (\cV, \cE_{\ML}(\brx^n))$ with vertex set $\cV$ and edge set $\cE_{\ML}(\brx^n)$. Given $P \in \cD(\cT^p, \theta)$, we are interested in the error event
\begin{equation}
\cA_P(n) := \{\cE_{\ML}(\brx^n) \neq \cE_{P}\} , \label{eq:AnDef}
\end{equation}
where the edge set $\cE_{\ML}(\brx^n)$, corresponding to the tree model returned by the ML estimator, is \emph{not} same as the edge set $\cE_{P}$ corresponding to the true graphical model $P$.

The following proposition shows that for the error event $\cA_P(n)$ is same as the event that the true tree distribution $P$ is not correctly estimated.
\begin{proposition} \label{prop:AnEquivalence}
	When $P \in \cD(\cT^p, \theta)$, and the learning algorithm has knowledge of $\theta$, then $\cA_P(n) = \{P_{\ML}(\brx^n) \neq P\}$.
\end{proposition} 
\begin{IEEEproof}
	See Appendix~\ref{app:AnEquivalence}.
\end{IEEEproof}
We remark that for general tree models, estimating the \emph{structure} of the tree is \emph{not} same as estimating the underlying distribution $P$ (as the parameters still have to be estimated). However, for our model, the properties P1 and P2 of the graphical model imply that the edge set uniquely characterizes the probability distribution.

\section{Maximum Likelihood Estimation and Error Exponent Analysis} \label{sec:ee}
We first study the algorithm for learning the ML tree distribution $P_{\ML}(\brx^n)$ given a set of $n$ samples $\brx^n$ drawn i.i.d. from a tree distribution $P \in \cD(\cX^p,\cT^p)$. Define the \emph{type} of $\brx^n$ to be the empirical distribution
\begin{equation}
\widehat{P}_{\brx^n}(\brx) := \frac{1}{n} \sum_{k=1}^n \one\{\brx_k = \brx\},~~ \brx \in \cX^p , \label{eq:Type_xn}
\end{equation}
where $\one\{\cdot\}$ denotes the indicator function. For notational convenience, in the rest of the paper we will denote the
empirical distribution $\widehat{P}_{\brx^n}$ by $\widehat{P}$. Combining \eqref{eq:P_ML_v1} and \eqref{eq:Type_xn}, we observe that
\begin{equation}
P_{\ML}(\brx^n) = \argmax_{Q \in \cD(\cT^p, \theta)} \sum_{\brx \in \cX^p} \widehat{P}(\brx) \log Q(\brx). \label{eq:P_ML_v2}
\end{equation}
Define the KL divergence, or relative entropy, between distributions $Q_1$ and $Q_2$ over alphabet $\cX^p$ as
\begin{equation*}
D(Q_1 \| Q_2) := \sum_{\brx \in \cX^p} Q_1(\brx) \log \frac{Q_1(\brx)}{Q_2(\brx)} . 
\end{equation*}
Then $D(Q_1 \| Q_2)$ can be equivalently expressed as
\begin{equation}
D(Q_1 \| Q_2) = -H(Q_1) - \sum_{\brx \in \cX^p} Q_1(\brx) \log Q_2(\brx) , \label{eq:KLdivergence}
\end{equation}
where $H(Q_1) :=- \sum_{\brx \in \cX^p} Q_1(\brx) \log Q_1(\brx)$ is the entropy of $Q_1$. From \eqref{eq:P_ML_v2} and \eqref{eq:KLdivergence}, it follows that $P_{\ML}(\brx^n)$ can be equivalently expressed as
\begin{equation}
P_{\ML}(\brx^n) = \argmin_{Q \in \cD(\cT^p, \theta)} D(\widehat{P} \| Q). \label{eq:P_ML_v3}
\end{equation}
The minimization over $Q$ in \eqref{eq:P_ML_v3} is known as the \emph{reverse I-projection}~\cite{Csiszar03_IP} of $\widehat{P}$ onto $\cD(\cT^p, \theta)$, the set of tree distributions satisfying properties P1 and P2 (parametrized by $\theta$). 

Let $\widehat{P}_{i,j}$ denote the marginal of $\widehat{P}$ on the pair of nodes $(i,j)$, with $i \neq j$, and define $\widehat{A}_{i,j}$ as
\begin{equation}
\widehat{A}_{i,j} := \widehat{P}_{i,j}(0,0) + \widehat{P}_{i,j}(1,1) . \label{eq:aij_Def}
\end{equation}
The following theorem shows that $P_{\ML}(\brx^n)$ can be efficiently computed using an MWST  algorithm, such as Prim's algorithm~\cite{AlgoBook09}, where the weight of the edge between nodes $i$ and $j$ is equal to $\widehat{A}_{i,j}$.

\begin{theorem} \label{thm:SimplifiedML}
	We have
	\begin{equation}
	P_{\ML}(\brx^n) = \argmax_{Q \in \cD(\cT^p, \theta)} \sum_{\{i,j\} \in \cE_Q} \widehat{A}_{i,j} , \label{eq:SimplifiedML}
	\end{equation}
	where $\cE_Q$ denotes the edge set of the tree distribution $Q$. Equivalently, $\cE_{\ML}(\brx^n)$ is the edge set of the MWST over a complete weighted graph where the weight of the edge $\{i,j\}$ is $\widehat{A}_{i,j}$.   
\end{theorem}
\begin{IEEEproof}
	See Appendix~\ref{app:SimplifiedML}.
\end{IEEEproof}
Note that the simplified rule for finding $\cE_{\ML}(\brx^n)$, given by Thm.~\ref{thm:SimplifiedML}, does not require explicit knowledge of $\theta$ (see Prop.~\ref{prop:AnEquivalence}), but merely exploits the fact that $\theta$ lies in the interval $(0,0.5)$.


\subsection{Error Exponent using Maximum Likelihood Estimation} \label{sec:ErrExpAnalysis}
Given $n$ samples, $\brx^n$, drawn i.i.d.\ from the distribution $P \in \cD(\cT^p, \theta)$, the error event $\cA_P(n)$, given by \eqref{eq:AnDef}, occurs when the ML estimator fails to correctly learn the edge set $\cE_{P}$. The \emph{error exponent} (also called the \emph{inaccuracy rate})~\cite{Tan11IT,Kester86}, captures the exponential decay of the error probability with the number of samples, and is formally defined as
\begin{equation}
K_P := \lim_{n \to \infty} -\frac{1}{n} \log \bbP\left(\cA_P(n)\right) , \label{eq:ErrExpDef}
\end{equation}  
where the limit was shown to exist by Tan, Anandkumar, Tong, and Willsky~\cite{Tan11IT}. Here, we assume that the number of nodes $p$ are fixed, while the number of samples $n$  drawn from $\cD(\cT^p, \theta)$  tends to infinity. We provide an exact explicit characterization for $K_P$ in Thm.~\ref{thm:ErrExpVal}.
\begin{theorem} \label{thm:ErrExpVal}
	For  $P \in \cD(\cT^p, \theta)$, we have
	\begin{equation}
	K_P = -\log\left(1 - \theta\big(1-\sqrt{4\theta(1-\theta)}\big)\right). \label{eq:ErrExpVal}
	\end{equation}
\end{theorem}
\begin{IEEEproof}
	See Appendix~\ref{app:ErrExpVal}.
\end{IEEEproof}
Note that the error exponent $K_P$ is independent of the number of nodes $p$ and the edge set $\cE_P$, and depends only on the parameter $\theta$.  

\subsection{Comparison with the classical Chow-Liu algorithm}
For scenarios where the   tree learning algorithm is \emph{not} aware of any additional tree property (such as properties P1 and P2 in our system model), an elegant solution to learning a general tree was presented by Chow and Liu in~\cite{ChowLiu68}. In particular, they showed that the edge set obtained using the Chow-Liu algorithm, denoted $\cE_{\CL}(\brx^n)$, is equal to the edge set of the MWST over a complete weighted graph where the weight of the edge $\{i,j\}$ is $I(\widehat{P}_{i,j})$, where $I(\widehat{P}_{i,j})$ denotes the \emph{empirical} mutual information between the pair of nodes $\{i,j\}$, 
\begin{equation*}
I(\widehat{P}_{i,j}) := \sum_{(x_i,x_j)\in \cX^2} \widehat{P}_{i,j}(x_i,x_j) \log\frac{\widehat{P}_{i,j}(x_i,x_j)}{\widehat{P}_{i}(x_i)\widehat{P}_{j}(x_j)} .
\end{equation*}

The paper by Tan, Anandkumar, Tong, and Willsky~\cite{Tan11IT} extended this line of work, and characterized the error exponent obtained using the Chow-Liu algorithm. Let $P$ be a tree-structured graphical model over 3 nodes with edge set $\cE_{P} = \left\{\{1,2\},\{2,3\}\right\}$, and define
\begin{align}
Q_*^{\CL} &:= \argmin_{Q \in \cP(\cX^3)} \left\{D(Q\|P) : I(Q_{1,2}) \le I(Q_{1,3})\right\}, \label{eq:QCL_star}\\
Q_{**}^{\CL} &:= \argmin_{Q \in \cP(\cX^3)} \left\{D(Q\|P) : I(Q_{2,3}) \le I(Q_{1,3})\right\}. \label{eq:QCL_starstar}
\end{align}
Then, the error exponent using the Chow-Liu algorithm, denoted $K_P^{\CL}$, can be expressed as~\cite{Tan11IT}
\begin{equation}
K_P^{\CL} = \min\left\{D(Q_*^{\CL} \| P) , \, D(Q_{**}^{\CL} \| P)\right\}.
\end{equation}
The following proposition shows that for $P \in \cD(\cT^3,\theta)$, the Chow-Liu error exponent is equal to the error exponent given by~\eqref{eq:ErrExpVal}.
\begin{proposition} \label{prop:CL_ErrExpVal}
	For $P \in \cD(\cT^3,\theta)$ with $\cE_{P} = \left\{\{1,2\},\{2,3\}\right\}$, we have
	\begin{equation}
	K_P^{\CL} = -\log\left(1 - \theta\big(1-\sqrt{4\theta(1-\theta)}\big)\right) = K_P. \label{eq:CL_ErrExpVal}
	\end{equation}
\end{proposition} 
\begin{IEEEproof}
	See Appendix~\ref{app:CL_ErrExpVal}.
\end{IEEEproof} 
From the {\em correlation decay property} for tree models with $p>3$ nodes having uniform marginals over a binary alphabet~\cite[Lem.~A.2]{Nikolakakis19Predictive}, and the fact that the dominant error event\footnote{We say that the event (more precisely, the sequence of events) $\cB_1(n)$ is {\em dominant} among a finite set of events $\{\cB_i(n)\}_{i=1}^k$ if $E_1=\min\{E_i: 1\le i\le k\}$ where   $E_i:=  \lim_{n\to\infty}-\frac{1}{n}\log\bbP(\cB_i(n))$  is the exponent of the probability of $\cB_i(n)$.} is the   in the learning problem occurs at various $3$-node sub-trees corresponding to nodes $\{i,j,k\} \subset \cV$ satisfying $\left\{\{i,j\}, \{j,k\}\right\} \subset \cE_P$ (see Appendix~\ref{app:ErrExpVal}), it follows that $K_P^{\CL} = K_P$ for general $P \in \cD(\cT^p,\theta)$. This observation that $K_P^{\CL} = K_P$ implies that from the error exponent perspective, somewhat surprisingly, there is no advantage in knowing that the tree-structured graphical model satisfies properties P1 and P2. However, we show via numerical simulations in Fig.~\ref{Fig:3nodesThetaPoint14}$(b)$ in Section~\ref{sec:3node} that when the sample size is extremely small, knowledge that the graphical model satisfies P1 and P2 yields smaller error probabilities over the vanilla Chow-Liu procedure. We also provide an intuitive reason for why this is the case in Sec.~\ref{sec:3node}.

\subsection{Comparison with related work}
Compared to the   model in Sec.~\ref{sec:SystemModel}, a more general tree model $P$ over a binary alphabet was analyzed by Bresler and Karzand in~\cite{BreslerKarzand18}. They assumed that the marginals of $P$ are uniform, but allowed for different correlations along the edges in $\cE_P$, i.e., for $\{i,j\} \in \cE_{P}$ it was assumed that $P_{i,j}(0,1) = P_{i,j}(1,0) = \theta_{i,j}/2$ where $\eta_1 \le |1 - 2\,\theta_{i,j}| \le \eta_2$ for some $0<\eta_1\le\eta_2<1$. Thus, our model  is a special case of that  in~\cite{BreslerKarzand18}, with $\theta_{i,j} = \theta < 0.5$ for all $\{i,j\} \in \cE_{P}$.

The result in \cite[Sec.~7.2]{BreslerKarzand18} can be specialized to our system model  to provide a non-asymptotic  upper bound on the error probability $\bbP\left(\cA_P(n)\right)$ as follows
\begin{equation}
\bbP\left(\cA_P(n)\right) \le 2 p^2 \exp\big(-n K_P^{\BK}\big), \label{eq:BK_BoundOnErrProb}
\end{equation}
where $K_P^{\BK}$ denotes the Bresler-Karzand exponent. This exponent, when specialized to our system model in which side information in the form of P1 and P2 is assumed, can be expressed as follows~\cite[Sec.~7.2]{BreslerKarzand18}
\begin{equation}
K_P^{\BK} := \frac{\theta \left(1 - 2 \theta \right)^2}{8}. \label{eq:BK_exp}
\end{equation} 

The following proposition compares $K_P^{\BK}$ with the {\em optimal} or {\em true} error exponent $K_P$.
\begin{proposition} \label{prop:BK_CompareExponent}
	For any $\theta \in (0,0.5)$ and $P \in \cD(\cT^p, \theta)$, we have
	\begin{equation}
	K_P^{\BK} < \frac{K_P}{3}.
	\end{equation}
\end{proposition}
\begin{IEEEproof}
	See Appendix~\ref{app:BK_CompareExponent}.
\end{IEEEproof}
Compared to asymptotic characterizations of the error probability (e.g., in Sec.~\ref{sec:ExactAsymptotics} to follow), the upper bound in~\eqref{eq:BK_BoundOnErrProb} has the advantage that it holds for all finite sample size $n\ge 1$. On the other hand, Prop.~\ref{prop:BK_CompareExponent} implies that the bound 
given by~\eqref{eq:BK_BoundOnErrProb} is rather loose asymptotically.

\section{Strong Large Deviations: Exact Asymptotics} \label{sec:ExactAsymptotics}
In the previous section, we showed that the error probability for learning a tree model $P \in \cD(\cT^p, \theta)$ decays exponentially with the number of samples $n$, and gave an explicit characterization of the error exponent in Thm.~\ref{thm:ErrExpVal}. 
In this section, we provide an exact and explicit characterization of the sub-exponential prefactor, resulting in a refined approximation for the error probability. Numerical results presented in Sec.~\ref{sec:NumericalResults} show that the resulting approximation provides a good fit for the empirical error probability obtained via Monte-Carlo simulations, even for relatively small values of $n$ (in the hundreds for a $10$-node tree).

The mainstay of our analysis is a strong large deviations theorem~\cite{BlackwellHodges59,BahadurRao60} that provides an asymptotic expansion of the logarithm of the probability of rare events of the form $\{\sum_{i=1}^n U_i \ge n \alpha\}$ for i.i.d. random variables $U_i$, $1\le i \le n$, where $\alpha$ is strictly larger than the mean of $U_1$. Under certain conditions, the asymptotic expansion is of the form $\bbP\left(\sum_{i=1}^n U_i \ge n \alpha\right) = \exp\left(-n \Lambda(\alpha) - (1/2)\log n + \gamma(\alpha) + o(1)\right)$, where $\Lambda(\cdot)$ is the large deviations function and $\gamma(\cdot)$ is a real-valued function. If we define $g_{\alpha}(n) := \exp\left(-n \Lambda(\alpha) - (1/2)\log n + \gamma(\alpha)\right)$, then $g_{\alpha}(n)$ is an asymptotically \emph{exact} approximation for $\bbP\left(\sum_{i=1}^n U_i \ge n \alpha\right)$ in the following sense:
\begin{equation*}
\lim_{n \to \infty} \frac{\bbP\left(\sum_{i=1}^n U_i \ge n \alpha\right)}{g_{\alpha}(n)} = 1 ~~\iff~~  \lim_{n \to \infty} \log \left( \frac{\bbP\left(\sum_{i=1}^n U_i \ge n \alpha\right)}{g_{\alpha}(n)} \right) = 0.
\end{equation*}
In contrast to strong large deviations, the (ordinary) large deviations analysis~\cite{HollanderBook00} only approximates $\bbP\left(\sum_{i=1}^n U_i \ge n \alpha\right)$ with the function $h_{\alpha}(n) :=  \exp\left(-n \Lambda(\alpha)\right)$, and hence
\begin{equation*}
\lim_{n \to \infty} \frac{1}{n} \log \left( \frac{\bbP\left(\sum_{i=1}^n U_i \ge n \alpha\right)}{h_{\alpha}(n)} \right) = 0.
\end{equation*}

We will approximate the error probability $\bbP\left(\cA_P(n)\right)$ with an explicitly defined function $g_P(n)$ that not only satisfies $\bbP\left(\cA_P(n)\right) = g_P(n)\big(1+o(1)\big)$, but also satisfies the sharper relation $\bbP\left(\cA_P(n)\right) = g_P(n)\big(1+o(n^{-1})\big)$. From a given $P \in \cD(\cT^p, \theta)$, we know from Appendix~\ref{app:ErrExpVal} that the error is dominated by events of the form $\big\{ \widehat{A}_{i,k} \ge \widehat{A}_{i,j} \big\}$, where $\left\{\{i,j\}, \{j,k\} \right\} \subset \cE_{P}$. The following lemma gives exact asymptotics for learning the graphical model of a $3$-node tree.

\begin{lemma} \label{lem:ExactAsymp3nodes}
	Let $P \in \cD(\cT^3, \theta)$ and $\cE_{P} = \left\{\{1,2\},\{2,3\}\right\}$. Define 	
	\begin{align}
	\tilde{f}(n) &:= \frac{\exp(-n K_P)}{\sqrt{2 \pi \sigma^2 n}} \left[1 + \frac{1 - 3\sigma^2}{8 \sigma^2 n} \right] , \label{eq:tilde_fn_Def}\\
	f(n) &:= \frac{\tilde{f}(n)}{1-z} \left[1 - \frac{z(1+z)}{2 (1-z)^2 \sigma^2 n} \right] , \label{eq:fn_Def}\\
	z &:= \sqrt{\frac{\theta}{1-\theta}} , \label{eq:z_Def}
	\end{align}
	where the exponent $K_P$ is given by \eqref{eq:ErrExpVal} and $\sigma^2 =  \theta \sqrt{4 \theta (1-\theta)}\, \exp (K_P)$. 
	Then, we have
	\begin{align}
	\bbP(\widehat{A}_{1,3} = \widehat{A}_{1,2}) \, &= \, \tilde{f}(n)\left( 1 + o(n^{-1})\right),\label{eq:ExactAsymp3nodes_eq_v1}\\
	\bbP\big(\widehat{A}_{1,3} \ge \widehat{A}_{1,2}\big) \, &= \, f(n) \left(1 + o(n^{-1})\right). \label{eq:ExactAsymp3nodes_ge_v1}
	\end{align} 	
\end{lemma}
\begin{IEEEproof}
	See Appendix~\ref{app:ExactAsymp3nodes_v1}.
\end{IEEEproof}
Note that when $P \in \cD(\cT^3, \theta)$ with $\cE_{P} = \left\{\{1,2\},\{2,3\}\right\}$, the event $\big\{ \widehat{A}_{2,3} > \widehat{A}_{1,2} = \widehat{A}_{1,3}\big\}$ does not guarantee an error in learning the tree structure. This is because it is still possible that $\cE_{\ML}(\brx^n) = \cE_{P}$, if the MWST  algorithm breaks ties  uniformly at random, and chooses the edge $\{1,2\}$ over $\{1,3\}$. In this case, given that event $\big\{ \widehat{A}_{2,3} > \widehat{A}_{1,2} = \widehat{A}_{1,3}\big\}$ has occurred, the probability of error in learning $\cE_P$ is equal to $1/2$. Similarly, the probability of error given that $\big\{ \widehat{A}_{2,3} = \widehat{A}_{1,2} = \widehat{A}_{1,3}\big\}$ is $2/3$. This observation, regarding randomly breaking ties in the MWST algorithm, can be used to obtain a sharp estimate for the error probability $\bbP(\cA_P(n))$ even for relatively small number of samples $n$. We remark that in the context of transmission of information over noisy channels, a similar idea using tie-breaking was employed in~\cite[Thm.~1]{Haim18} to provide an improved upper bound on the error probability. This technique, employing tie-breaking while estimating the error probability, assumes importance in scenarios where the probability of ties has roughly the same order as the total error probability.

The above learning algorithm can be made conservative via post-processing, whereby an error is declared if there exists an edge $\{i_1,i_2\} \notin \cE_{\ML}(\brx^n)$ such that $\widehat{A}_{i_1,i_2} = \min_{\{i,j\} \in \cE_{\ML}(\brx^n)} \widehat{A}_{i,j}$. For a $3$-node tree with $\cE_{P} = \left\{\{1,2\},\{2,3\}\right\}$, the error asymptotics with this conservative rule is given, using \eqref{eq:ExactAsymp3nodes_ge_v1}, as follows\footnote{The exponent of $\bbP\big(\big\{\widehat{A}_{1,3} \ge \widehat{A}_{1,2}\big\} \cap \big\{\widehat{A}_{1,3} \ge \widehat{A}_{2,3}\big\}\big)$ (i.e., the exponential rate of the decrease of this probability to zero) is strictly larger than the exponent of $\bbP\big(\widehat{A}_{1,3} \ge \widehat{A}_{1,2}\big)$ and $\bbP\big(\widehat{A}_{1,3} \ge \widehat{A}_{2,3}\big)$ (see Appendix~\ref{app:ExactAsymp3nodes}).}
\begin{equation*}
\bbP(\cA_P(n)) = \bbP\big(\{\widehat{A}_{1,3} \ge \widehat{A}_{1,2}\} \cup \{\widehat{A}_{1,3} \ge \widehat{A}_{2,3}\}\big) \, = \, 2f(n) \left(1 + o(n^{-1})\right).
\end{equation*} 	
The following proposition shows that the above error asymptotics is \emph{strictly} worse than the error asymptotics obtained for a tie-breaking MWST algorithm.

\begin{proposition} \label{prop:ExactAsymp3nodes}
	When $P \in \cD(\cT^3, \theta)$, and ties are randomly broken in an MWST algorithm, then
	\begin{equation}
	\bbP(\cA_P(n)) = \big(2 f(n) - \tilde{f}(n)\big)\big(1+o(n^{-1})\big) , \label{eq:ExactAsymp3nodes}
	\end{equation}
	where $\tilde{f}(n)$ and $f(n)$ are given by \eqref{eq:tilde_fn_Def} and \eqref{eq:fn_Def}, respectively.
\end{proposition}
\begin{IEEEproof}
	See Appendix~\ref{app:ExactAsymp3nodes}.
\end{IEEEproof}

The following theorem generalizes the result in Prop.~\ref{prop:ExactAsymp3nodes} to $p \ge 3$ nodes.
\begin{theorem} \label{thm:ExactAsymptotics}
	For $P \in \cD(\cT^p,\theta)$, let $T_P = (\cV, \cE_P)$ be the tree graph of the graphical model $P$. For $1 \le i \le p$, let $d_i$ denote the degree of node $i$ in $T_P$, and define
	\begin{equation}
	\dP := \sum_{i=1}^p \frac{d_i (d_i - 1)}{2} . \label{eq:dP_Def}
	\end{equation}
	When ties are randomly broken in an MWST algorithm, then we have
	\begin{equation}
	\bbP(\cA_P(n)) = \dP \big(2 f(n) - \tilde{f}(n)\big)\big(1+o(n^{-1})\big). \label{eq:ExactAsymptotics}
	\end{equation}
\end{theorem}
\begin{IEEEproof}
	See Appendix~\ref{app:ExactAsymptotics}.
\end{IEEEproof}
We remark that the key step to generalize Prop.~\ref{prop:ExactAsymp3nodes} to Thm.~\ref{thm:ExactAsymptotics} is to incorporate the multiplicative factor $\dP$, which accounts for the number of $3$-node sub-trees of $\cT_P$ that contribute to dominant errors~\cite[App.~A.1]{Nikolakakis19NonParametric} in the ML learning algorithm. Note that while $f(n)$ and $\tilde{f}(n)$ do not depend on the particular choice of $P$, the multiplicative factor $\dP$ depends on the tree structure $T_P$ via the degrees of the respective nodes in $T_P$. Thm.~\ref{thm:ExactAsymptotics} provides an explicit function $g_P(n) = \dP \big(2 f(n) - \tilde{f}(n)\big)$, that closely approximates the error probability as $\bbP\big(\cA_P(n)\big) = g_P(n)\big(1+o(n^{-1})\big)$. We also remark that it is not difficult to extend Thm.~\ref{thm:ExactAsymptotics} to the case in which the homogeneity property P2 does not hold (each $3$-node sub-tree would have its own $\theta_{i,j}$ and $\theta_{j,k}$), but then the result would be more cumbersome to state.

\section{Extending Exact Asymptotics to Noisy Samples Setting} \label{sec:ExactAsymptotics_NS}
This section considers the scenario where the observed samples are noise-corrupted versions of the samples generated from the underlying tree-structured graphical model. This setup, where we only have access to noisy samples, has practical applications, including scenarios where measurement errors are introduced in systems with limited precision. 

We consider a hidden Markov random field with hidden layer $\brx = [x_1, \ldots, x_p] \sim P \in \cD(\cT^p,\theta)$, and observed (noisy) sample $\bry = [y_1, \ldots, y_p] \sim P^{(q)}$, where $\bry$ is the output when each component of $\brx$ is passed through a memoryless binary symmetric channel (BSC) with crossover probability $0 \le q < 0.5$. The output distribution $P^{(q)}$ is expressed as follows
\begin{equation}
P^{(q)}(\bry) = \sum_{\brx \in \cX^p} q^{\delta_{\brx,\bry}} (1-q)^{p - \delta_{\brx,\bry}} P(\brx), ~~~~ \bry \in \cY^p = \{0,1\}^p , \label{eq:Dist_Pq}
\end{equation}
where $\delta_{\brx,\bry} := \sum_{k=1}^n \one\{ x_k \ne y_k\}$ denotes the \emph{Hamming distance} between $\brx$ and $\bry$. Note that $P^{(q)} = P$ for $q=0$. 

In the  noisy sample setting, the problem is to learn the edge set $\cE_{P}$ of the underlying tree model $P$, using $n$ noisy samples $\bry^n := \{\bry_1,\ldots,\bry_n\}$, where the distribution of each noisy sample is given by \eqref{eq:Dist_Pq}. As in the noiseless case, the side information to the tree learning algorithm is the knowledge that the underlying graphical model $P$ satisfies   properties P1 and P2.

Given $n$ noisy samples $\bry^n$, the empirical distribution of $\bry^n$, denoted $\widehat{P}^{(q)}_{\bry^n}$, is given by 
\begin{equation}
\widehat{P}^{(q)}_{\bry^n}(\bry) := \frac{1}{n} \sum_{k=1}^n \one\{\bry_k = \bry\},~~ \bry \in \cY^p . \label{eq:Type_yn}
\end{equation}
For notational convenience, we will denote the empirical distribution $\widehat{P}^{(q)}_{\bry^n}$ by $\widehat{P}^{(q)}$. Let $\widehat{P}^{(q)}_{i,j}$ denote the marginal of $\widehat{P}^{(q)}$ on the pair of nodes $(i,j)$  and define 
\begin{equation}
\widehat{A}^{(q)}_{i,j} := \widehat{P}^{(q)}_{i,j}(0,0) + \widehat{P}^{(q)}_{i,j}(1,1) . \label{eq:aijq_Def}
\end{equation}
For a given graphical model $P$ with edge set $\cE_{P}$, we denote the estimated edge set (using $n$ noisy samples $\bry^n$) as $\widehat{\cE}^{(q)}(\bry^n)$. We use a learning algorithm that returns $\widehat{\cE}^{(q)}(\bry^n)$ as the edge set of an MWST over a complete weighted graph where the weight of $\{i,j\}$ is equal to $\widehat{A}^{(q)}_{i,j}$. The following proposition shows that this algorithm yields  the ML estimate of $\cE_{P}$ when $P \in \cD(\cT^3, \theta)$.

\begin{proposition} \label{prop:EstimateEdgeSet_NS}
	When $P \in \cD(\cT^3, \theta)$, an ML estimate of $\cE_{P}$, using $n$ noisy samples $\bry^n$, is obtained as the edge set of an MWST over a complete weighted graph where the weight of $\{i,j\}$ is equal to $\widehat{A}^{(q)}_{i,j}$ given by \eqref{eq:aijq_Def}.
\end{proposition}
\begin{IEEEproof}
	See Appendix~\ref{app:EstimateEdgeSet_NS}.
\end{IEEEproof}

\subsection{Error Exponent: Noisy Samples}
Given $n$ noisy samples, $\bry^n$, drawn i.i.d. from $P^{(q)}$, we want to analyze the error probability $\bbP\big( \hat{\cE}^{(q)}(\bry^n) \neq \cE_{P}\big)$. Towards this, we first quantify the error exponent, denoted $K_P^{(q)}$, associated with the error probability $\bbP\big( \hat{\cE}^{(q)}(\bry^n) \neq \cE_{P}\big)$. Formally, we have
\begin{equation}
K_P^{(q)} := \lim_{n \to \infty} -\frac{1}{n} \log \bbP\big( \hat{\cE}^{(q)}(\bry^n) \neq \cE_{P}\big) . \label{eq:ErrExpDef_NS}
\end{equation}  
The following theorem provides an exact explicit characterization of $K_P^{(q)}$. 

\begin{theorem} \label{thm:KPq_Val}
	We have
	\begin{equation}
	K_P^{(q)} = -\log\left(1 - 4\Big(\frac{\beta_1 + \beta_2}{2} - \sqrt{\beta_1 \beta_2} \Big)\right) , \label{eq:KPq_Val}
	\end{equation}
	where
	\begin{align}
	\beta_1 &:= \left( (1-q)^3 + q^3 \right)\frac{\theta(1-\theta)}{2} + q(1-q) \bigg(\frac{1 - \theta(1-\theta)}{2}\bigg), \label{eq:beta1}\\*
	\beta_2 &:= \left( (1-q)^3 + q^3 \right)\frac{\theta^2}{2} + q(1-q) \bigg(\frac{1 - \theta^2}{2}\bigg) . \label{eq:beta2}
	\end{align}
\end{theorem}
\begin{IEEEproof}
	See Appendix~\ref{app:KPq_Val}.
\end{IEEEproof}
Comparing Thm.~\ref{thm:ErrExpVal} and Thm.~\ref{thm:KPq_Val}, we observe that $K_P^{(q)} = K_P$ when $q=0$. Thus, the error exponent using noisy samples, given by \eqref{eq:KPq_Val}, generalizes the exponent for the noiseless setting. Prop.~\ref{prop:EstimateEdgeSet_NS}, together with the fact that $3$-node error events are dominant~\cite[Sec.~4, App.~A.1]{Nikolakakis19NonParametric}, implies that $K_P^{(q)}$ in~\eqref{eq:KPq_Val} is the {\em optimal}  error exponent for learning trees with noisy samples.

\subsection{Exact Asymptotics: Noisy Samples} 
We now proceed with the main result of this section, where we present the exact asymptotics for the error probability for the scenario where we only have access to noise-corrupted samples for learning the underlying tree structure. This result generalizes the exact asymptotics using noiseless samples presented in Sec.~\ref{sec:ExactAsymptotics}. 

\begin{theorem} \label{thm:ExactAsymp_NS}
	Let $P \in \cD(\cT^p,\theta)$, and let $P^{(q)}$ be given by~\eqref{eq:Dist_Pq}. When $\bry^n$ are $n$ i.i.d. samples distributed according to $P^{(q)}$, then we have
	\begin{equation}
	\bbP\big( \hat{\cE}^{(q)}(\bry^n) \neq \cE_{P}\big) \,=\, \dP \big(2 f^{(q)}(n) - \tilde{f}^{(q)}(n)\big)\big(1 + o(n^{-1})\big), \label{eq:ExactAsymp_NS}
	\end{equation}
	where $\dP$ is given by~\eqref{eq:dP_Def}, and
	\begin{align}
	\tilde{f}^{(q)}(n) &:=  \frac{\exp\big(-n K_P^{(q)}\big)}{\sqrt{2 \pi \mu_2 n}} \left[1 + \frac{1 - 3\mu_2}{8 \mu_2 n} \right] , \label{eq:tilde_fq_Def}\\
	f^{(q)}(n) &:= \frac{\tilde{f}^{(q)}(n)}{1-z} \left[1 - \frac{z(1+z)}{2 (1-z)^2 \mu_2 n}\right],	\label{eq:fq_Def}\\
	\mu_2 &:= 4\sqrt{\beta_1 \beta_2}\, \exp\big(K_P^{(q)}\big) ,\label{eq:Mu2_NS} \\
	z &:= \sqrt{\frac{\beta_2}{\beta_1}} , \label{eq:z_NS_Def}
	\end{align}
	where $K_P^{(q)}$, $\beta_1$, and $\beta_2$ are given by \eqref{eq:KPq_Val}, \eqref{eq:beta1}, and \eqref{eq:beta2}, respectively. 
\end{theorem}
\begin{IEEEproof}
	See Appendix~\ref{app:ExactAsymp_NS}.
\end{IEEEproof}
Similar to the noiseless case, the functions $f^{(q)}(n)$ and $\tilde{f}^{(q)}(n)$ do not depend on the particular choice of $P$, while the multiplicative factor $\dP$ depends on the tree structure $T_P$ via the degree of respective nodes in $T_P$. Thm.~\ref{thm:ExactAsymp_NS} provides us with an explicitly defined function $g_P^{(q)}(n) = \dP \big(2 f^{(q)}(n) - \tilde{f}^{(q)}(n)\big)$  that closely approximates the error probability as $\bbP\big(\hat{\cE}^{(q)}(\bry^n) \neq \cE_{P}\big) = g_P^{(q)}(n) \big(1 + o(n^{-1})\big)$. Note that when $q=0$, we have $f^{(0)}(n) = f(n)$ and $\tilde{f}^{(0)}(n) = \tilde{f}(n)$, and thus the exact error asymptotics in Thm.~\ref{thm:ExactAsymp_NS} generalizes the result \eqref{eq:ExactAsymptotics} for the noiseless setting.

\subsection{Related Result for Noisy Samples}
In \cite{Nikolakakis19AISTATS,Nikolakakis19Predictive}, Nikolakakis, Kalogerias, and Sarwate extended Bresler and Karzand's result~\cite{BreslerKarzand18} to provide bounds on the number of \emph{noisy} samples required to achieve a given target probability for learning an Ising tree model. Similar to~\cite{BreslerKarzand18}, the model in \cite{Nikolakakis19AISTATS} is more general compared to that described in Sec.~\ref{sec:SystemModel}, as it allows different correlations along the edges of the underlying tree. As our model in Sec.~\ref{sec:SystemModel} is a special case of the model in~\cite{Nikolakakis19AISTATS}, the results in~\cite{Nikolakakis19AISTATS} can be specialized to our system model. In particular, \cite[Thm.~1]{Nikolakakis19AISTATS} (or equivalently \cite[Thm.~3.1]{Nikolakakis19Predictive}) can be adapted to provide a non-asymptotic  upper bound on $\bbP\big( \hat{\cE}^{(q)}(\bry^n) \neq \cE_{P}\big)$ as follows:
\begin{equation}
\bbP\big( \hat{\cE}^{(q)}(\bry^n) \neq \cE_{P}\big) \le 2 p^2 \exp\big(-n K_P^{\NKS(q)}\big). \label{eq:NKS_BoundOnErrProb}
\end{equation}
Here, $K_P^{\NKS(q)}$ denotes the Nikolakakis-Kalogerias-Sarwate exponent, given by~\cite[Eq.~(14)]{Nikolakakis19AISTATS} and restated here as follows:
\begin{equation}
K_P^{\NKS(q)} := \frac{(1-2q)^4 \theta^2 (1-2\theta)^2}{8 \left(1 - (1-2q)^4 (1-2\theta)\right)}. \label{eq:NKS_exp}
\end{equation} 
In Sec.~\ref{sec:comparison}, we numerically compare $K_P^{\NKS(q)}$ in~\eqref{eq:NKS_exp} to $K_P^{(q)}$ in~\eqref{eq:KPq_Val}  and show that the exponent $K_P^{\NKS(q)}$ is much smaller than the true exponent $K_P^{(q)}$. This implies that the upper bound in~\eqref{eq:NKS_BoundOnErrProb} is rather loose in relation to the exact asymptotics given by~\eqref{eq:ExactAsymp_NS}.

\section{Numerical Results} \label{sec:NumericalResults}
This section presents numerical results, and illustrates that our theoretical results for the noiseless and noisy sample scenarios are in keen agreement with the empirical observations. 

\subsection{Comparison of different exponents} \label{sec:comparison}
\begin{figure}[t]
	\centering
	\includegraphics[width=1.05\textwidth, angle=0]{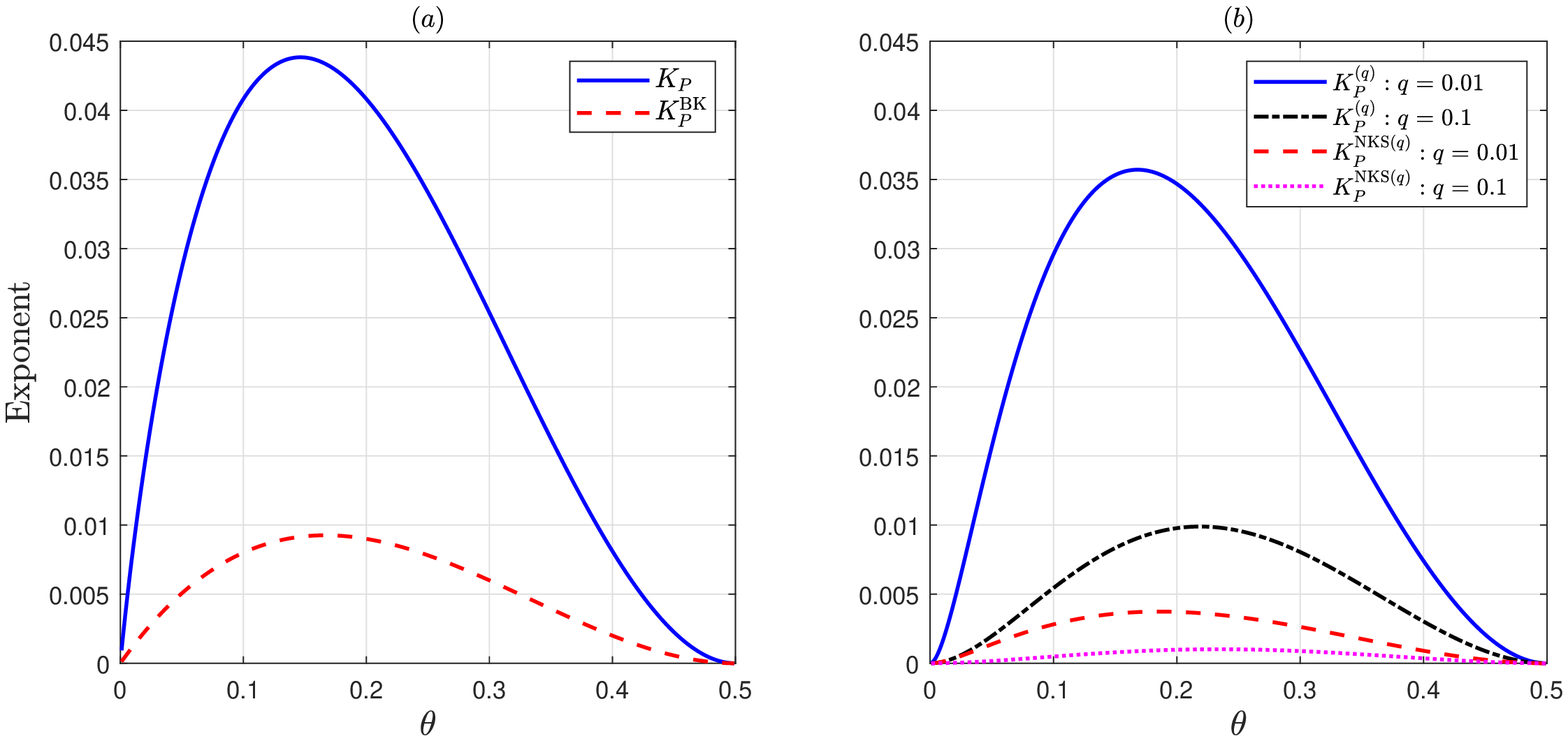}
	\caption{$(a)$ compares the error exponent $K_P$ with $K_P^{\BK}$, the Bresler-Karzand exponent~\cite[Sec.~7.2]{BreslerKarzand18}, for the noiseless setting, while $(b)$ compares the error exponent $K_P^{(q)}$ with $K_P^{\NKS(q)}$, the Nikolakakis-Kalogerias-Sarwate exponent~\cite[Thm.~1]{Nikolakakis19AISTATS},\cite[Thm.~3.1]{Nikolakakis19Predictive}, for the noisy samples setting.}
	\label{Fig:BK_NKS_Exponent}
\end{figure}

It is well known that the error exponent captures the asymptotics behavior of the error probability~\cite{CoverBook06}, and so formulations with imprecise exponents are expected to provide inaccurate approximations for error probabilities when the sample size is relatively large.  Fig.~\ref{Fig:BK_NKS_Exponent}$(a)$ compares the error exponent $K_P$ in~\eqref{eq:ErrExpVal} for the noiseless scenario, with the corresponding exponent $K_P^{\BK}$ in~\eqref{eq:BK_exp} based on the work by Bresler-Karzand~\cite[Sec.~7.2]{BreslerKarzand18}. As stated in Prop.~\ref{prop:BK_CompareExponent} and also shown in Fig.~\ref{Fig:BK_NKS_Exponent},  $K_P^{\BK}$ is significantly smaller than the true exponent $K_P$, and hence the upper bound on the error probability given by~\eqref{eq:BK_BoundOnErrProb} can only provide a weak estimate of the error probability. Fig.~\ref{Fig:BK_NKS_Exponent}$(b)$ compares the exponent using noisy samples $K_P^{(q)}$ in~\eqref{eq:KPq_Val} with the corresponding exponent $K_P^{\NKS(q)}$ in~\eqref{eq:NKS_exp}, based on the work by Nikolakakis-Kalogerias-Sarwate~\cite[Thm.~1]{Nikolakakis19AISTATS},\cite[Thm.~3.1]{Nikolakakis19Predictive}, for BSC crossover probabilities $q=0.01$ and $q=0.1$. As expected, the exponents in Fig.~\ref{Fig:BK_NKS_Exponent}$(b)$ are smaller than exponents for the noiseless setting in Fig.~\ref{Fig:BK_NKS_Exponent}$(a)$, and it is observed that the exponent decreases with an increase in $q$. Fig.~\ref{Fig:BK_NKS_Exponent}$(b)$ demonstrates a large gap between $K_P^{(q)}$ and $K_P^{\NKS(q)}$, implying that the upper bound given by~\eqref{eq:NKS_BoundOnErrProb} is rather loose.

\subsection{Comparison of theoretical and simulation results: $3$-node tree} \label{sec:3node}
\begin{figure}[t]
	\centering
	\includegraphics[width=1.05\textwidth, angle=0]{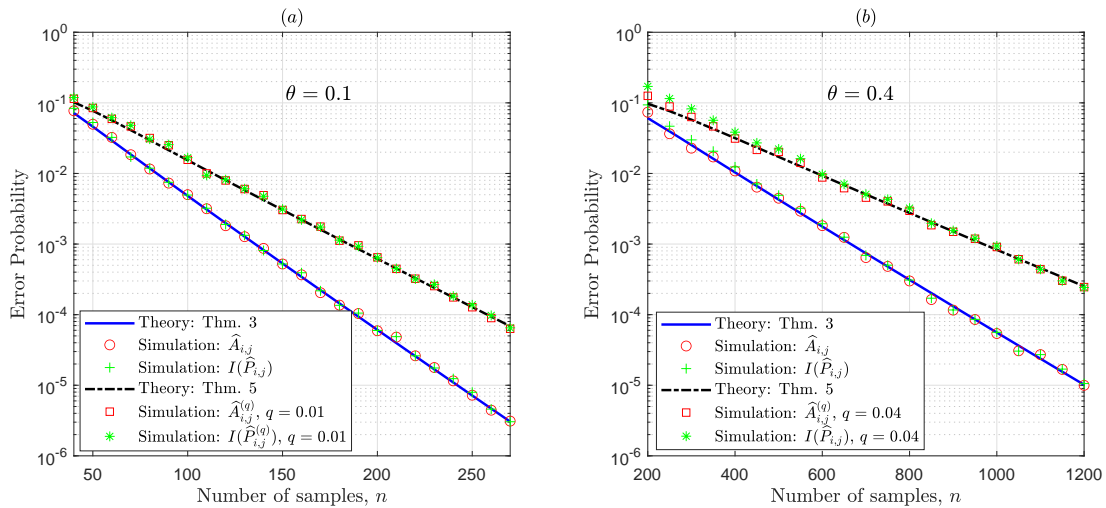}
	\caption{Comparison of the theoretical error asymptotics for the noiseless and noisy sample setting for a $3$-node tree ($p=3$) via Thm.~\ref{thm:ExactAsymptotics} and Thm.~\ref{thm:ExactAsymp_NS}, respectively, with corresponding simulation results. Simulation results labeled with $\widehat{A}_{i,j}$ (resp.\ $I(\widehat{P}_{i,j})$) imply that the estimated tree is the output of an MWST algorithm whose input is a complete graph with edge $\{i,j\}$ weighted with $\widehat{A}_{i,j}$ (resp.\ $I(\widehat{P}_{i,j})$), and so on for $\widehat{A}_{i,j}^{(q)}$ and $I(\widehat{P}_{i,j}^{(q)})$.}
	\label{Fig:3nodesThetaPoint14}
\end{figure}

Fig.~\ref{Fig:3nodesThetaPoint14} compares the exact asymptotics for learning a $3$-node tree for the noiseless and noisy sample setting with corresponding simulation results. The theoretical result for the noiseless samples setting ($q=0$) is obtained using Thm.~\ref{thm:ExactAsymptotics} with error probability approximated by $\dP \big(2 f(n) - \tilde{f}(n)\big)$ in \eqref{eq:ExactAsymptotics}\ (i.e., we ignore the multiplicative factor $(1+o(n^{-1}) )$). The theoretical result for the noisy samples setting ($q>0$) is obtained using Thm.~\ref{thm:ExactAsymp_NS} with error probability approximated by $\dP \big(2 f^{(q)}(n) - \tilde{f}^{(q)}(n)\big)$ in~\eqref{eq:ExactAsymp_NS}. For the noiseless samples case, the simulation results are obtained using synthetically generated data samples with distribution $P$ satisfying properties P1 and P2 (Sec.~\ref{sec:SystemModel}). For the noisy samples setting, the generated samples have distribution $P^{(q)}$ in~\eqref{eq:Dist_Pq}. For each parameter setting, the number of iterations for obtaining the simulated error probability was chosen to ensure that at least $200$ errors  occurred. Simulation results labeled with $\widehat{A}_{i,j}$ (resp.\ $I(\widehat{P}_{i,j})$) imply that the estimated tree is the output of an MWST algorithm whose input is a complete graph with edge $\{i,j\}$ weighted with $\widehat{A}_{i,j}$ (resp.\ $I(\widehat{P}_{i,j})$), and so on for $\widehat{A}_{i,j}^{(q)}$ and $I\big(\widehat{P}_{i,j}^{(q)}\big)$. When we use the mutual information quantities $I(\widehat{P}_{i,j})$ and $I\big(\widehat{P}_{i,j}^{(q)}\big)$, we are running the vanilla Chow-Liu algorithm~\cite{ChowLiu68}, i.e., we are not leveraging the side information that P1 and P2 hold.

Fig.~\ref{Fig:3nodesThetaPoint14}$(a)$ and $(b)$ compare the theoretical and simulated results for $\theta=0.1$ and $\theta=0.4$, respectively.  These figures demonstrate  that the theoretical estimates of the error probabilities, given by Thm.~\ref{thm:ExactAsymptotics} and Thm.~\ref{thm:ExactAsymp_NS}, closely match  the simulation results. In comparison, the upper bounds on the error probability given by~\eqref{eq:BK_BoundOnErrProb} and~\eqref{eq:NKS_BoundOnErrProb} evaluate to more than $1$ for the parameters chosen for Fig.~\ref{Fig:3nodesThetaPoint14}, and hence are not plotted.
The Chow-Liu algorithm~\cite{ChowLiu68} is seen to perform almost similarly to the ML algorithm.  However,  the former is marginally worse than the latter when $n$ is small for $\theta=0.4$, exhibiting the benefit of side information. Roughly speaking, this is because   errors in the Chow-Liu algorithm for a $3$-node tree satisfying P1 and P2 arise when the empirical binary entropy of the estimated parameter of the non-edge, say $H(\widehat{\theta}_{1,3})$, is not larger than that of a true edge, say $H(\widehat{\theta}_{1,2})$. This is dominated by the event $\{ H(\widehat{\theta}_{1,2})=H(\widehat{\theta}_{1,3})\}$. By the symmetry of the binary entropy function around $1/2$, we see that that is equivalent to   $\{\widehat{\theta}_{1,2} =\widehat{\theta}_{1,3} \}\cup\{\widehat{\theta}_{1,2} =1-\widehat{\theta}_{1,3} \}$. In contrast, the ML algorithm with side information using $\widehat{A}_{i,j}$ or $\widehat{A}_{i,j}^{(q)}$ only errs when $\{\widehat{\theta}_{1,2} =\widehat{\theta}_{1,3} \}$ holds. Hence, there is a slight benefit of the ML algorithm over the Chow-Liu algorithm especially when $n$ is small and $\theta$ is close to $1/2$.

\subsection{Comparison of theoretical and simulation results: $10$-node trees}

\begin{figure}[t]
	\hspace{-10mm}\includegraphics[width=1.15\textwidth, angle=0]{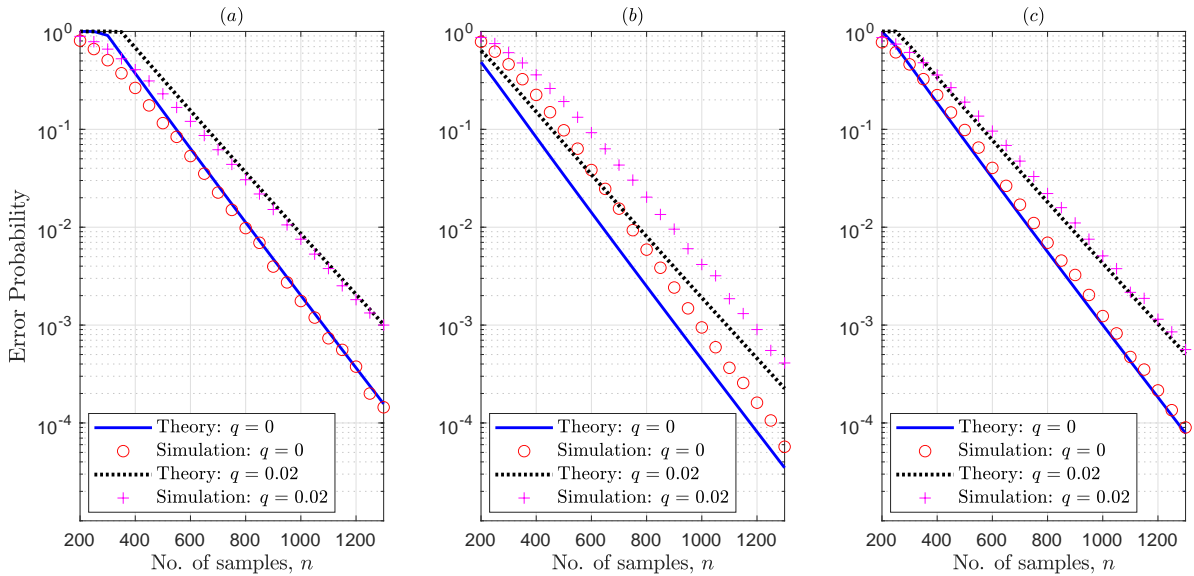}
	\caption{Comparison of the theoretical asymptotics with simulation results for the noiseless ($q=0$) and noisy sample setting ($q=0.02$) for $10$-node trees with $\theta=0.4$. Figures $(a)$, $(b)$, and $(c)$  correspond to \emph{star} ($\dP=36$), \emph{Markov chain} ($\dP=8$), and \emph{hybrid} ($\dP=18$) tree structures, respectively.}
	\label{Fig:10nodesThetaPoint4}
\end{figure}
Fig.~\ref{Fig:10nodesThetaPoint4} compares the theoretical and simulated error asymptotics with $\theta=0.4$ for $10$-node trees whose structures are $(a)$ star, $(b)$ Markov chain, and $(c)$ hybrid, where we follow the definitions of these tree structures as given in~\cite{Tan10TSP}. Also, the extremal properties of the star and Markov chain tree structure were highlighted in~\cite{Tan10TSP}. Similar to the observations in~\cite{Tan10TSP}, we note from Thm.~\ref{thm:ExactAsymptotics} and Thm.~\ref{thm:ExactAsymp_NS}, that for a $p$-node tree, the error probability is asymptotically maximal (resp.\ minimal) for a star (resp.\ Markov chain) tree structure, due to the corresponding structure having a maximal (resp.\ minimal) value of $\dP$ in~\eqref{eq:dP_Def} (see Appendix~\ref{app:StarMC}). For the simulation results, the estimated tree is the output of an MWST algorithm whose input is a complete graph with edge $\{i,j\}$ weighted with $\widehat{A}_{i,j}$  (resp.\ $\widehat{A}_{i,j}^{(q)}$) for the noiseless (resp.\ noisy) samples setting. Again, the simulated error probability is obtained by averaging over a number of iterations such that at least  $200$ errors occurred. Fig.~\ref{Fig:10nodesThetaPoint4} shows an overall agreement between the theoretical and simulation results, even for moderate values of $n$. In contrast,  the upper bounds on the error probability given by~\eqref{eq:BK_BoundOnErrProb} and~\eqref{eq:NKS_BoundOnErrProb} evaluate to more than $1$ for the parameters chosen for Fig.~\ref{Fig:10nodesThetaPoint4}, and hence are not plotted.  An interesting observation one can make from Fig.~\ref{Fig:10nodesThetaPoint4}$(b)$ is that for the Markov chain, the simulated error probability is generally higher than the theoretical prediction. This is because the theoretical analysis only captures {\em dominant} error events; however, for the chain, there are many non-dominant error events that contribute to the simulated error probability and this effect is more pronounced at small sample sizes. In contrast,  for the star, {\em all} ``single-edge error events'' (i.e., error events in which the true and estimated edge sets differ by one edge) are dominant, and hence the theoretical results are close to simulation results for   sample sizes $n\ge 600$. The behavior of the hybrid tree naturally lies in between those of the extremal structures.

\section{Reflections}\label{sec:reflect}
This paper has taken a first step in refining estimates of the error probability in learning graphical models. We have taken a strong large deviations approach to compute the exact asymptotics for learning trees given noiseless and noisy samples. For the noiseless and noisy cases respectively, we have significantly improved on the error exponents derived by Bresler-Karzand~\cite{BreslerKarzand18} and Nikolakakis-Kalogerias-Sarwate~\cite{Nikolakakis19AISTATS}. The theoretical results show keen agreement with numerical simulations at relatively small sample sizes. We believe the analytical techniques developed here are novel in statistical learning  and may be broadly applicable to other learning  problems  with discrete solutions such as ranking and feature subset selection. 

There are a few promising avenues for future research. What we have done thus far pertains to the {\em low-dimensional setting} in which $p$ is fixed and $n$ grows. Because of the asymptotic tools used, our results cannot be directly extended to the more practically relevant  {\em high-dimensional} setting in which $p$ grows simultaneously with $n$. Another direction of research would be to use the analytical tools herein to analyze the probability of error for learning other classes of graphical models such as random graphs~\cite{Anandkumar12}, latent tree   models~\cite{Choi11}, or more general Ising models~\cite{Bresler15STOC}. Finally, we would like to explore if the suite of strong large deviation techniques employed here can be used to sharpen upper~\cite{Dasarathy16AISTATS} and lower bounds~\cite{Scarlett17AISTATS} for the {\em active} learning of graphical models or error estimates of other machine learning tasks.

\appendices
\section{Proof of Proposition~\ref{prop:AnEquivalence}} \label{app:AnEquivalence}
We have to show $\{\cE_{\ML}(\brx^n) \neq \cE_{P}\} = \{P_{\ML}(\brx^n) \neq P\}$ in order to prove Prop.~\ref{prop:AnEquivalence}. 

\begin{itemize}
	\item $\{P_{\ML}(\brx^n) \neq P\} \subset \{\cE_{\ML}(\brx^n) \neq \cE_{P}\}$: From \eqref{eq:TreeFactorizationP}, P1, and P2 (in Sec.~\ref{sec:SystemModel}), it follows that the underlying distribution can be explicitly stated based on the knowledge of the edge set. Therefore, the correct determination of the edge set implies the correct determination of the underlying distribution, and so $\{\cE_{\ML}(\brx^n) = \cE_{P}\} \subset \{P_{\ML}(\brx^n) = P\}$, and the claim is proved by contraposition.
	\item $\{\cE_{\ML}(\brx^n) \neq \cE_{P}\} \subset \{P_{\ML}(\brx^n) \neq P\}$: As $|\cE_{\ML}(\brx^n)| = |\cE_{P}| = p-1$, the condition $\cE_{\ML}(\brx^n) \neq \cE_{P}$ implies that there exists an edge $\{i,j\} \in \cE_{P}$, such that $\{i,j\} \notin \cE_{\ML}(\brx^n)$. Let $P_{\ML(i,j)}$ and $P_{i,j}$ denote the marginals on edge $\{i,j\}$ for $P_{\ML}(\brx^n)$ and $P$, respectively. As $\{i,j\} \notin \cE_{\ML}(\brx^n)$, combining properties P1 and P2 (Sec.~\ref{sec:SystemModel}) and the \emph{correlation decay property} for tree models with uniform marginal distribution on each node over a binary alphabet~\cite[Lem.~A.2]{Nikolakakis19Predictive}, it follows that $P_{\ML(i,j)}(0,0) + P_{\ML(i,j)}(1,1) < 1-\theta = P_{i,j}(0,0) + P_{i,j}(1,1)$. This implies $P_{\ML(i,j)} \neq P_{i,j}$, and hence $\{P_{\ML}(\brx^n) \neq P\}$.
\end{itemize}
\qed

\section{Proof of Theorem~\ref{thm:SimplifiedML}} \label{app:SimplifiedML}
From \eqref{eq:P_ML_v2}, we have
\begin{align*}
P_{\ML}(\brx^n) &= \argmax_{Q \in \cD(\cT^p, \theta)} \sum_{\brx \in \cX^p} \widehat{P}(\brx) \log Q(\brx) , \\
&\overset{(a)}{=} \argmax_{Q \in \cD(\cT^p, \theta)} \sum_{\brx \in \cX^p} \widehat{P}(\brx) \bigg(\log Q_1(x_1) + \sum_{i=2}^p \log Q_{i|\pa(i)}\left(x_i | x_{\pa(i)}\right) \bigg),  \\
&\overset{(b)}{=} \argmax_{Q \in \cD(\cT^p, \theta)} \sum_{i=2}^p \bigg( \sum_{\brx \in \cX^p} \widehat{P}(\brx)  \log Q_{i|\pa(i)}\left(x_i | x_{\pa(i)}\right) \bigg), \\ 
&= \argmax_{Q \in \cD(\cT^p, \theta)} \sum_{i=2}^p \bigg( \sum_{(x_i, x_{\pa(i)}) \in \cX^2} \widehat{P}_{i,\pa(i)}(x_i, x_{\pa(i)})  \log Q_{i|\pa(i)}\left(x_i | x_{\pa(i)}\right) \bigg), \\
&\overset{(c)}{=} \argmax_{Q \in \cD(\cT^p, \theta)} \sum_{i=2}^p \left( \widehat{A}_{i,\pa(i)} \log (1-\theta) + \big(1-\widehat{A}_{i,\pa(i)}\big) \log \theta \right), \\
&= \argmax_{Q \in \cD(\cT^p, \theta)} \sum_{i=2}^p  \widehat{A}_{i,\pa(i)} \log \Big(\frac{1-\theta}{\theta}\Big), \\
&\overset{(d)}{=} \argmax_{Q \in \cD(\cT^p, \theta)} \sum_{i=2}^p  \widehat{A}_{i,\pa(i)} = \argmax_{Q \in \cD(\cT^p, \theta)} \sum_{\{i,j\} \in \cE_Q} \widehat{A}_{i,j} ,
\end{align*}
where $(a)$ follows from \eqref{eq:TreeFactorizationP},\, $(b)$ follows from the fact that $Q_1(0) = Q_1(1) = 0.5$ (property P1),\, $(c)$ follows from \eqref{eq:aij_Def} and the facts that $Q_{i|\pa(i)}(0|0) = Q_{i|\pa(i)}(1|1) = 1-\theta$, and $Q_{i|\pa(i)}(0|1) = Q_{i|\pa(i)}(1|0) = \theta$ (property P2),\, and $(d)$ follows from the fact that $1-\theta > \theta$.
\qed

\section{Proof of Theorem~\ref{thm:ErrExpVal}} \label{app:ErrExpVal}
We will first prove the following lemma, which will be applied to prove Thm~\ref{thm:ErrExpVal}.
\begin{lemma} \label{lem:ErrExp3nodes}
	Let $\tilde{P} \in \cP(\cX^3)$ be defined as $\tilde{P}(x_1,x_2,x_3) = \tilde{P}_1(x_1) \tilde{P}_{2|1}(x_2|x_1) \tilde{P}_{3|2}(x_3|x_2)$ with uniformly distributed $\tilde{P}_1$, and $\tilde{P}_{2|1}(0|1)
=\tilde{P}_{2|1}(1|0) = \theta_1 < 0.5$, and $\tilde{P}_{3|2}(0|1)
=\tilde{P}_{3|2}(1|0) = \theta_3 < 0.5$. Let $\brx^n$ denote the $n$ i.i.d. samples drawn from $\tilde{P}$, let $\widehat{P}$ denote the type of $\brx^n$, and let $\widehat{A}_{i,j}$ be given by \eqref{eq:aij_Def} for $1 \le i < j \le 3$. Then, we have
\begin{equation}
\lim_{n \to \infty} -\frac{1}{n} \log \bbP\big(\widehat{A}_{1,3} \ge \widehat{A}_{1,2}\big) = -\log\left(1 - \theta_3\big(1-\sqrt{4\theta_1(1-\theta_1)}\big)\right). \label{eq:ErrExp3nodes_v1}
\end{equation}
\end{lemma}
\begin{IEEEproof}
	From Sanov's theorem~\cite[Thm.~11.4.1]{CoverBook06}, it follows that
	\begin{align}
 &\lim_{n \to \infty}	-\frac{1}{n} \log \bbP\big(\widehat{A}_{1,3} \ge \widehat{A}_{1,2}\big) \nonumber\\*
 &\qquad = \min_{Q \in \cP(\cX^3)} \left\{D(Q \| \tilde{P}) : Q_{1,3}(0,0) + Q_{1,3}(1,1) \ge Q_{1,2}(0,0) + Q_{1,2}(1,1) \right\}, \label{eq:ErrExp3nodes_v2}
	\end{align} 
	where $Q_{i,j}$ denotes the marginal of $Q$ for the pair of nodes $(i,j)$. The constraint $Q_{1,3}(0,0) + Q_{1,3}(1,1) \ge Q_{1,2}(0,0) + Q_{1,2}(1,1)$ is equivalent to the following constraint
	\begin{equation}
	Q(0,1,0)+Q(1,0,1) \ge Q(0,0,1)+Q(1,1,0). \label{eq:SanovConstraint}
	\end{equation}
	Let $Q^*$ denote that $Q \in \cP(\cX^3)$ which satisfies \eqref{eq:SanovConstraint} and minimizes $D(Q\|\tilde{P})$. Let $g$ be the map $g : \cX^3 \to \bbR$ defined as follows
	\begin{equation} \label{eq:Define_g}
	\begin{aligned}
	&g(0,1,0) = g(1,0,1) = 1,~~~~g(0,0,1) = g(1,1,0) = -1, \\
	&g(0,0,0) = g(1,1,1) = g(0,1,1) = g(1,0,0) = 0.
	\end{aligned}
	\end{equation}
	Then, by using a Lagrange multiplier, $Q^*$ can be obtained to be the \emph{tilted distribution}~\cite{Moulin17},
	\begin{equation}
	Q^*(\brx) = \frac{\tilde{P}(\brx) \exp\left(\lambda g(\brx)\right)}{\sum_{\bry \in \cX^3}\tilde{P}(\bry) \exp\left(\lambda g(\bry)\right)}, ~~\brx \in \cX^3, \label{eq:TiltedDistQ}
	\end{equation}
	where $\lambda$ is chosen to satisfy \eqref{eq:SanovConstraint}, and is given by 
	\begin{equation}
	\lambda = \frac{1}{2} \log\bigg(\frac{\tilde{P}(0,0,1)}{\tilde{P}(0,1,0)}\bigg). \label{eq:lambdaVal}
	\end{equation}
	Now, for any two real numbers $a>0$ and $b>0$, we define
	\begin{equation}
	\Delta(a,b) := \frac{a+b}{2} - \sqrt{ab}, \label{eq:TriangleDef}
	\end{equation} 
	and hence $\Delta(a,b)$ is the difference between the arithmetic and geometric means of $a$ and $b$, and $\Delta(a,b) \ge 0$ with equality if and only if $a=b$. Now, using \eqref{eq:ErrExp3nodes_v2} we have
	\begin{align}
	\lim_{n \to \infty} -\frac{1}{n} \log \bbP\big(\widehat{A}_{1,3} \ge \widehat{A}_{1,2}\big) &= D(Q^* \| \tilde{P}) , \nonumber \\
	&= -\log\left(1 - 4 \Delta\big(\tilde{P}(0,0,1),\,\tilde{P}(0,1,0)\big) \right) , \label{eq:ErrExp3nodes_v3} \\
	&= -\log\left(1 - \theta_3\big(1-\sqrt{4\theta_1(1-\theta_1)}\big)\right), \nonumber
	\end{align}
	where \eqref{eq:ErrExp3nodes_v3} follows using \eqref{eq:TiltedDistQ} and \eqref{eq:lambdaVal}.
\end{IEEEproof}
We apply Lem.~\ref{lem:ErrExp3nodes} to $\tilde{P} \in \cD(\cT^3, \theta)$ to obtain the following proposition.
\begin{proposition} \label{prop:ErrExp3nodes}
	For $\tilde{P} \in \cD(\cT^3, \theta)$, we have
	\begin{equation*}
	K_{\tilde{P}} = -\log\left(1 - \theta\big(1-\sqrt{4\theta(1-\theta)}\big)\right) .
	\end{equation*}
\end{proposition} 
\begin{IEEEproof}
	Without loss of generality, assume that $\cE_{\tilde{P}} = \left\{ \{1,2\}, \{2,3\} \right\}$. It follows from Thm.~\ref{thm:SimplifiedML} that when $n$ i.i.d. samples drawn from $\tilde{P}$ are used for learning the tree structure, the error event $\{\cE_{\ML}(\brx^n) \neq \cE_{\tilde{P}}\}$ occurs when $\widehat{A}_{1,3} \ge \widehat{A}_{1,2}$ or $\widehat{A}_{1,3} \ge \widehat{A}_{2,3}$. From symmetry, we have $\bbP\big(\widehat{A}_{1,3} \ge \widehat{A}_{1,2}\big) =  \bbP\big(\widehat{A}_{1,3} \ge \widehat{A}_{1,2}\big)$, and hence
	\begin{equation}
	K_{\tilde{P}} = \lim_{n \to \infty} -\frac{1}{n} \log \bbP\big(\widehat{A}_{1,3} \ge \widehat{A}_{1,2}\big) = -\log\left(1 - \theta \big(1-\sqrt{4\theta(1-\theta)}\big)\right) , \label{eq:KP_tilde}
	\end{equation}
	where the last equality follows from Lem.~\ref{lem:ErrExp3nodes}.
\end{IEEEproof}
For a general $p$-node tree distribution $P \in \cD(\cT^p, \theta)$, it follows from \cite[App.~A.1]{Nikolakakis19NonParametric} that the dominant error event in the learning problem occurs at various $3$-node sub-trees corresponding to the tree distribution $P$, and takes one the following forms: $\big\{\widehat{A}_{i,k} \ge \widehat{A}_{i,j}\big\}$ or $\big\{\widehat{A}_{i,k} \ge \widehat{A}_{j,k}\big\}$, where $\left\{ \{i,j\}, \{j,k\} \right\} \subset \cE_{P}$. This observation is also related to the correlation decay property for tree models with uniform marginals over binary alphabet~\cite[Lem.~A.2]{Nikolakakis19Predictive}. For a given $p$, the number of such $3$-node sub-trees is fixed, and hence it follows from Prop.~\ref{prop:ErrExp3nodes} that the error exponent in learning $P$ is given by $K_P=-\log \big(1 - \theta\big(1-\sqrt{4\theta(1-\theta)}\big)\big)$.

Before concluding this appendix, we highlight the intuition regarding the dominant error occurring at $3$-node sub-trees mentioned above. Towards this, consider the example where $p=4$, $P \in \cD(\cT^4, \theta)$ with $\cE_{P} = \left\{\{1,2\},\{2,3\},\{3,4\}\right\}$. In this case, it follows from Lem.~\ref{lem:ErrExp3nodes} that 
\begin{equation}
\lim_{n \to \infty} -\frac{1}{n} \log \bbP\big(\widehat{A}_{1,3} \ge \widehat{A}_{1,2}\big) = -\log\left(1 - \theta\big(1-\sqrt{4\theta(1-\theta)}\big)\right). \label{eq:ErrExp3nodes_v4}
\end{equation}
We will characterize exactly the exponent corresponding to the error event $\big\{\widehat{A}_{1,4} \ge \widehat{A}_{1,2}\big\}$, using Lem.~\ref{lem:ErrExp3nodes}, and show that it is strictly higher than the exponent in~\eqref{eq:ErrExp3nodes_v4}. Note that the probability $\mathbb{P}\big(\widehat{A}_{1,4} \ge \widehat{A}_{1,2}\big)$ can be characterized by analyzing the marginal distribution $P_{1,2,4}(x_1,x_2,x_4)$. Due to the Markov property, we have $P_{4|1,2}(x_4 | x_1, x_2) = P_{4|2}(x_4 | x_2)$ where $P_{4|2}(0 | 1) = P_{4|2}(1 | 0) = 2\theta(1-\theta)$. Therefore, applying Lem.~\ref{lem:ErrExp3nodes} and taking $\theta_1 = \theta$, $\theta_3 = 2\theta(1-\theta)$, we obtain
\begin{equation}
\lim_{n \to \infty} -\frac{1}{n} \log \bbP\big(\widehat{A}_{1,4} \ge \widehat{A}_{1,2}\big) = -\log\left(1 - \theta_3\big(1-\sqrt{4\theta(1-\theta)}\big)\right). \label{eq:ErrExp3nodes_v5}
\end{equation}
Similarly, it can be shown that the exponent corresponding to the error event $\big\{\widehat{A}_{1,4} \ge \widehat{A}_{2,3}\big\}$ is also given by the right side of \eqref{eq:ErrExp3nodes_v5}, which is a \emph{strictly} increasing function of $\theta_3$. Thus, the exponent in \eqref{eq:ErrExp3nodes_v5} is strictly greater than that in \eqref{eq:ErrExp3nodes_v4} because $\theta_3 = 2 \theta (1-\theta) > \theta$ for $0 < \theta < 0.5$, thereby showing that the error event $\big\{\widehat{A}_{1,4} \ge \widehat{A}_{1,2}\big\}$ (or $\big\{\widehat{A}_{1,4} \ge \widehat{A}_{2,3}\big\}$) does not dominate the overall expression for the error probability.
\qed

\section{Proof of Proposition~\ref{prop:CL_ErrExpVal}} \label{app:CL_ErrExpVal}
For $P \in \cD(\cT^3,\theta)$ with $\cE_{P} = \left\{\{1,2\},\{2,3\}\right\}$, we have $P_{1,2} = P_{2,3}$, and so comparing \eqref{eq:QCL_star} and \eqref{eq:QCL_starstar}, it follows that $D(Q_*^{\CL} \| P) = D(Q_{**}^{\CL} \| P)$, and hence
\begin{equation}
K_P^{\CL} = \min_{Q \in \cP(\cX^3)} \left\{D(Q \| P) : I(Q_{1,2}) \le I(Q_{1,3})\right\} . \label{eq:CL_EE_v1}
\end{equation}
Further, the marginals of $P$ are uniformly distributed, and it follows from symmetry that any distribution $Q$ that minimizes \eqref{eq:CL_EE_v1} also has uniform marginals. Now, if we define $\gamma_{1,2} := Q_{1,2}(0,1) + Q_{1,2}(1,0)$ and $\gamma_{1,3} := Q_{1,3}(0,1) + Q_{1,3}(1,0)$, then we have $I(Q_{1,2}) = 1 - H(\gamma_{1,2})$ and $I(Q_{1,3}) = 1 - H(\gamma_{1,3})$, where $H(\cdot)$ is the binary entropy function, and $Q$ has uniform marginals. As $H(\cdot)$ is symmetric about 0.5, the constraint $I(Q_{1,2}) \le I(Q_{1,3})$ is satisfied if and only if one of the following linear constraints is satisfied: 
\begin{enumerate}[(i)]
	\item $0.5 > \gamma_{1,2} \ge \gamma_{1,3}$.
	\item $0.5 > \gamma_{1,2}, \ \gamma_{1,3} \ge 1 - \gamma_{1,2}$.
	\item $0.5 < \gamma_{1,2} \le \gamma_{1,3}$.
	\item $0.5 < \gamma_{1,2}, \ \gamma_{1,3} \le 1 - \gamma_{1,2}$.
	\item $0.5 = \gamma_{1,2} = \gamma_{1,3}$.
\end{enumerate}
As $P_{1,2}(0,1)+P_{1,2}(1,0) = \theta < 0.5$ and $P_{1,3}(0,1)+P_{1,3}(1,0) = 2\theta(1-\theta) < 0.5$, it follows that the constraint on $Q$ that minimizes $D(Q\|P)$ is $\gamma_{1,3} \le \gamma_{1,2} < 0.5$, and hence \eqref{eq:CL_EE_v1} can be equivalent expressed as
\begin{equation}
K_P^{\CL} = \min_{Q \in \cP(\cX^3)} \left\{D(Q \| P) : Q_{1,3}(0,1)+Q_{1,3}(1,0) \le Q_{1,2}(0,1)+Q_{1,2}(1,0) < 0.5 \right\} . \label{eq:CL_EE_v2}
\end{equation}
Finally, comparing \eqref{eq:CL_EE_v2} with \eqref{eq:ErrExp3nodes_v2}, and applying \eqref{eq:KP_tilde}, we obtain \eqref{eq:CL_ErrExpVal}.
\qed

\section{Proof of Proposition~\ref{prop:BK_CompareExponent}} \label{app:BK_CompareExponent}
We define $\vartheta := \theta(1 - \sqrt{4 \theta (1-\theta)})$, and using \eqref{eq:ErrExpVal} we obtain
\begin{align}
K_P &= \log\left(\frac{1}{1 - \vartheta}\right) > \log(1 + \vartheta) , \nonumber \\
&\overset{(\mathrm{i})}{>} \frac{2 \vartheta}{2 + \vartheta} \, \overset{(\mathrm{ii})}{>}\, \frac{2 \vartheta}{2.5}  \,>\, \frac{3 \theta(1 - \sqrt{4 \theta (1-\theta)})}{4} , \label{eq:KP_LowerBound}
\end{align}
where $(\mathrm{i})$ follows by applying~\cite[Eq.~(3)]{Topsoe04}, and $(\mathrm{ii})$ follows because $\vartheta < 0.5$. On the other hand, using~\eqref{eq:BK_exp} we have
\begin{align}
K_P^{\BK} &= \frac{\theta(1 - 4\theta + 4 \theta^2)}{8} = \frac{\theta(1 - \sqrt{4\theta(1-\theta)})(1 + \sqrt{4\theta(1-\theta)})}{8}, \nonumber \\
&\overset{(\mathrm{iii})}{<}\frac{\theta(1 - \sqrt{4\theta(1-\theta)})}{4} \overset{(\mathrm{iv})}{<} \frac{K_P}{3} , \nonumber
\end{align}
where $(\mathrm{iii})$ follows because $\sqrt{4\theta(1-\theta)} < 1$, and $(\mathrm{iv})$ follows from~\eqref{eq:KP_LowerBound}.
\qed

\section{Proof of Lemma~\ref{lem:ExactAsymp3nodes}} \label{app:ExactAsymp3nodes_v1}
For $n$ i.i.d. discrete random variables $\{U_i\}_{i=1}^n$ taking integer values whose differences have greatest common divisor equal to $1$, Blackwell and Hodges~\cite{BlackwellHodges59} gave exact asymptotic expansions for the probabilities $\bbP\left(\sum_{i=1}^n U_i = n \alpha \right)$ and $\bbP\left(\sum_{i=1}^n U_i \ge n \alpha \right)$, under the condition that $\bbP\left(\sum_{i=1}^n U_i = n \alpha \right) > 0$ for every admissible $n$, and where $\alpha > \bbE[U_1]$. We will apply this result by appropriately defining $U_1$ and $\alpha$. We first prove \eqref{eq:ExactAsymp3nodes_eq_v1}. For $P \in \cD(\cT^3, \theta)$, each random sample belongs to the alphabet $\cX^3$, and we define $U_1$ as follows:
\begin{equation*}
U_1 = \begin{cases}
1   &\mathrm{~if~} \brx \in \left\{(0,1,0), (1,0,1) \right\}, \\
-1 &\mathrm{~if~} \brx \in \left\{(0,0,1), (1,1,0) \right\}, \\
0 &\mathrm{~if~} \brx \in \left\{(0,0,0), (1,1,1), (0,1,1), (1,0,0) \right\}.
\end{cases}
\end{equation*}
Then, we have
\begin{align*}
\big\{ \widehat{A}_{1,3} = \widehat{A}_{1,2} \big\} &= \big\{\widehat{P}(0,1,0) + \widehat{P}(1,0,1) - \widehat{P}(0,0,1) - \widehat{P}(1,1,0) = 0 \big\},\\
&= \bigg\{\sum_{i=1}^n U_i = 0 \bigg\}.
\end{align*}  
As $P \in \cD(\cT^3, \theta)$, $\cE_{P} = \left\{\{1,2\}, \{2,3\}\right\}$, and $0 < \theta < 0.5$, we have
\begin{equation*}
P(0,0,1) + P(1,1,0) = \theta(1-\theta) > \theta^2 = P(0,1,0) + P(1,0,1) .
\end{equation*}
It follows that if we choose $\alpha=0$, then we have $(\mathrm{i})$ $0 > \bbE[U_1]$, and $(\mathrm{ii})$ $\bbP\big(\sum_{i=1}^n U_i = 0 \big) > 0$ for every admissible $n$. Thus, the variables satisfy the required conditions for applying the exact asymptotic theorems presented in~\cite{BlackwellHodges59}. 

Note that $U_1$ takes values in the ternary alphabet $\cU =\{-1,0,1\}$, and we denote the probability distribution of $U_1$ as $Q(u)$, $u \in \cU$. Now, as $\alpha=0$, it is sufficient for us to consider the moment generating function of $U_1$ defined as
\begin{equation*}
\phi(t) := \bbE_Q\left[e^{t U_1}\right] = Q(-1)\,e^{-t} + Q(0) + Q(1)\,e^t,
\end{equation*}
and let $\tau$ be the value of $t$ which minimizes $\phi(t)$, i.e. $\tau = \argmin_{t\in\mathbb{R}}\phi(t)$. Then, $\tau$ is uniquely determined as the solution of $\phi^{\prime}(\tau) = 0$, which gives us 
\begin{align}
e^{\tau} &= \sqrt{\frac{Q(-1)}{Q(1)}} = \sqrt{\frac{1-\theta}{\theta}}, \label{eq:tau_val}\\
\phi(\tau) &= Q(0) + \sqrt{4 Q(-1) Q(1)} = (1-\theta) + \theta\sqrt{4 \theta (1-\theta)}= \exp(-K_P), \label{eq:PhiTau_and_KP}
\end{align}
where \eqref{eq:PhiTau_and_KP} follows from \eqref{eq:ErrExpVal}. Now, we define a random variable $V_1$ taking values in $\cU$ having an \emph{exponentially tilted distribution}~\cite{Moulin17}, denoted $\tilde{Q}$, and defined as follows
\begin{equation*}
\tilde{Q}(u) := \frac{Q(u) e^{\tau u}}{\phi(\tau)} , ~~~u \in \cU.
\end{equation*}
Note that $\bbE_{\tilde{Q}}[V_1] = \phi^{\prime}(\tau)/\phi(\tau) = 0$. Let $\sigma^2$, $\mu_3$, and $\mu_4$ denote the second, third, and fourth central moments of $V_1$, respectively. Then, we have
\begin{align*}
\sigma^2 &= \frac{\phi^{\prime \prime}(\tau)}{\phi(\tau)} = \frac{Q(-1)e^{-\tau} + Q(1) e^{\tau}}{\phi(\tau)} = \theta \sqrt{4 \theta (1-\theta)}\, \exp(K_P) , \\
\mu_3 &= \frac{\phi^{\prime\prime \prime}(\tau)}{\phi(\tau)} = \frac{\phi^{\prime}(\tau)}{\phi(\tau)} = 0 , \qquad \quad \mu_4 = \frac{\phi^{\prime\prime \prime\prime}(\tau)}{\phi(\tau)} = \frac{\phi^{\prime\prime}(\tau)}{\phi(\tau)} = \sigma^2 . \label{eq:Mu4Val}
\end{align*}
From the strong large deviations theorem~\cite[Thm.~3]{BlackwellHodges59}, we have
\begin{equation}
\bbP\bigg(\sum_{i=1}^n U_i = 0\bigg) \, = \, \frac{[\phi(\tau)]^n}{\sqrt{2 \pi \sigma^2 n}} \left[1 + \frac{1}{8n}\left(\frac{\mu_4}{\sigma^4} -3 -\frac{5 \mu_3^2}{3 \sigma^6} \right) \right]\left( 1 + o(n^{-1})\right). \label{eq:ExactAsymp_eq_BH}
\end{equation}
Using the fact that $\bbP\left(\sum_{i=1}^n U_i = 0\right) \,=\, \bbP\big(\widehat{A}_{1,3} = \widehat{A}_{1,2}\big)$, and substituting the values of $\phi(\tau)$, $\sigma^2$, $\mu_3$, and $\mu_4$, in \eqref{eq:ExactAsymp_eq_BH}, we obtain
\begin{equation}
\bbP \bigg(\sum_{i=1}^n U_i = 0\bigg) \,=\, \tilde{f}(n) \left(1 + o(n^{-1}) \right) \,=\, \bbP \big(\widehat{A}_{1,3} = \widehat{A}_{1,2}\big) , \label{eq:ExactAsymp3nodes_eq_v2}
\end{equation}
where $\tilde{f}(n)$ is given by \eqref{eq:tilde_fn_Def}, thereby completing the proof of \eqref{eq:ExactAsymp3nodes_eq_v1}.

Next, we proceed to prove \eqref{eq:ExactAsymp3nodes_ge_v1}. The strong large deviations theorem given in~\cite[Thm.~4]{BlackwellHodges59} states that the exact asymptotics of $\bbP\left(\sum_{i=1}^n U_i \ge 0\right)$ is given by
\begin{equation}
\bbP \bigg(\sum_{i=1}^n U_i \ge 0\bigg) \,=\, \frac{\tilde{f}(n)}{1-z} \left[1 - \frac{1}{2n}\left(\frac{(z \mu_3/\sigma^2) + z(1+z)/(1-z)}{(1-z) \sigma^2}\right)\right] \left(1 + o(n^{-1})\right) , \label{eq:ExactAsymp3nodes_ge_v2}
\end{equation}
where $z = e^{-\tau} = \sqrt{\theta/(1-\theta)}$. As $\mu_3 = 0$, the expression in \eqref{eq:ExactAsymp3nodes_ge_v2} simplifies to
\begin{equation}
\bbP \bigg(\sum_{i=1}^n U_i \ge 0 \bigg) \,=\, \frac{\tilde{f}(n)}{1-z} \left[1 - \frac{z(1+z)}{2 (1-z)^2 \sigma^2 n}\right] \left(1 + o(n^{-1})\right) , \label{eq:ExactAsymp3nodes_ge_v3}
\end{equation}
and the proof of \eqref{eq:ExactAsymp3nodes_ge_v1} is complete by using \eqref{eq:fn_Def}, \eqref{eq:ExactAsymp3nodes_ge_v3}, and noting that $\bbP\big(\widehat{A}_{1,3} \ge \widehat{A}_{1,2}\big) \,=\, \bbP\left(\sum_{i=1}^n U_i \ge 0\right)$.
\qed

\section{Proof of Proposition~\ref{prop:ExactAsymp3nodes}} \label{app:ExactAsymp3nodes}
For $P \in \cD(\cT^3, \theta)$, we assume without loss of generality that $\cE_{P} = \left\{\{1,2\}, \{2,3\}\right\}$. Here, if $\big\{\widehat{A}_{1,3} > \widehat{A}_{1,2}\big\}$ or $\big\{\widehat{A}_{1,3} > \widehat{A}_{2,3}\big\}$, an MWST algorithm will pick the incorrect edge $\{1,3\} \in \cE_{\ML}(\brx^n)$. Hence, $\big\{\cE_{\ML}(\brx^n) \neq \cE_{P}\big\}$ surely  occurs if  $\big\{\widehat{A}_{1,3} > \widehat{A}_{1,2}\big\}$ or $\big\{\widehat{A}_{1,3} > \widehat{A}_{2,3}\big\}$ occur. However, in the case of tie-breaking, we have the following scenarios:
\begin{itemize}
	\item $\cE_{\ML}(\brx^n) \neq \cE_{P}$ with probability $1/2$ when $\big\{\widehat{A}_{2,3} > \widehat{A}_{1,2} = \widehat{A}_{1,3}\big\}$ or $\big\{\widehat{A}_{1,2} > \widehat{A}_{2,3} = \widehat{A}_{1,3}\big\}$.
	\item $\cE_{\ML}(\brx^n) \neq \cE_{P}$ with probability $2/3$ when $\big\{\widehat{A}_{1,2} = \widehat{A}_{2,3} = \widehat{A}_{1,3}\big\}$.
\end{itemize} 

Therefore, the probability of error $\bbP(\cA_P(n))$ is given by
\begin{align}
\bbP(\cA_P(n)) &= \bbP\big(\big\{\widehat{A}_{1,3} > \widehat{A}_{1,2}\big\}  \cup \big\{\widehat{A}_{1,3} > \widehat{A}_{2,3}\big\} \big) + \frac{1}{2}\, \bbP\big(\widehat{A}_{2,3} > \widehat{A}_{1,2} = \widehat{A}_{1,3} \big) \nonumber \\
&\qquad + \frac{1}{2}\, \bbP\big(\widehat{A}_{1,2} > \widehat{A}_{2,3} = \widehat{A}_{1,3} \big) + \frac{2}{3}\, \bbP\big(\widehat{A}_{1,2} = \widehat{A}_{2,3} = \widehat{A}_{1,3} \big). \label{eq:ExAsy3node_v1}
\end{align}
The individual probability components in \eqref{eq:ExAsy3node_v1} satisfy the following relations:
\begin{align}
\bbP\big(\big\{\widehat{A}_{1,3} > \widehat{A}_{1,2}\big\}  \cup \big\{\widehat{A}_{1,3} > \widehat{A}_{2,3}\big\} \big) &= \bbP\big(\widehat{A}_{1,3} > \widehat{A}_{1,2}\big) + \bbP\big(\widehat{A}_{1,3} > \widehat{A}_{2,3}\big) \nonumber \\
&\qquad\qquad - \bbP\big(\big\{\widehat{A}_{1,3} > \widehat{A}_{1,2}\big\}  \cap \big\{\widehat{A}_{1,3} > \widehat{A}_{2,3}\big\} \big), \label{eq:ExAsy3node_P1} \\
\bbP\big(\widehat{A}_{1,2} > \widehat{A}_{2,3} \! =\!  \widehat{A}_{1,3} \big) &\! =\!   \bbP\big(\widehat{A}_{2,3} \! =\!  \widehat{A}_{1,3}\big)\!  -\!  \bbP\big(\big\{\widehat{A}_{2,3} \! =\!  \widehat{A}_{1,3}\big\} \cap \big\{\widehat{A}_{1,2} \le \widehat{A}_{1,3}\big\} \big) \label{eq:ExAsy3node_P2} \\
\bbP\big(\widehat{A}_{2,3} > \widehat{A}_{1,2} \! = \! \widehat{A}_{1,3} \big) &\! =\!  \bbP\big(\widehat{A}_{1,2} \! =\!  \widehat{A}_{1,3}\big) \! -\!  \bbP\big(\big\{\widehat{A}_{1,2}\!  =\!  \widehat{A}_{1,3}\big\} \cap \big\{\widehat{A}_{2,3} \le \widehat{A}_{1,3}\big\} \big). \label{eq:ExAsy3node_P3}
\end{align}
From \eqref{eq:ExactAsymp3nodes_eq_v1}, \eqref{eq:ExactAsymp3nodes_ge_v1}, and the symmetry in $\cE_P$, it follows that the exponents corresponding to probabilities $\bbP\big(\widehat{A}_{1,3} > \widehat{A}_{1,2}\big)$, $\bbP\big(\widehat{A}_{1,3} > \widehat{A}_{2,3}\big)$, $\bbP\big(\widehat{A}_{1,3} = \widehat{A}_{1,2}\big)$, and
$\bbP\big(\widehat{A}_{1,3} = \widehat{A}_{2,3}\big)$ are equal to $K_P$ defined in~\eqref{eq:ErrExpVal}. On the other hand, the exponents corresponding to the probabilities $\bbP\big(\big\{\widehat{A}_{1,3} > \widehat{A}_{1,2}\big\}  \cap \big\{\widehat{A}_{1,3} > \widehat{A}_{2,3}\big\} \big)$, $\bbP\big(\big\{\widehat{A}_{2,3} = \widehat{A}_{1,3}\big\} \cap \big\{\widehat{A}_{1,2} \le \widehat{A}_{1,3}\big\} \big)$, $\bbP\big(\big\{\widehat{A}_{1,2} = \widehat{A}_{1,3}\big\} \cap \big\{\widehat{A}_{2,3} \le \widehat{A}_{1,3}\big\} \big)$, and $\bbP\big(\widehat{A}_{1,2} = \widehat{A}_{2,3} = \widehat{A}_{1,3} \big)$ are strictly greater than $K_P$.\footnote{We have $\lim_{n \to \infty}-\frac{1}{n}\log \bbP\big(\{\widehat{A}_{1,3} \ge \widehat{A}_{1,2}\big\} \cap \{\widehat{A}_{1,3} \ge \widehat{A}_{2,3}\} \big) = -\log\big(1-\theta \big[2 - \theta - 3 \theta^{1/3} (1-\theta)^{2/3} \big]\big) > K_P$.}
Using \eqref{eq:ExAsy3node_v1}, \eqref{eq:ExAsy3node_P1}, \eqref{eq:ExAsy3node_P2}, and \eqref{eq:ExAsy3node_P3}, and collecting the terms with the smallest exponent $K_P$, we have
\begin{align}
\bbP(\cA_P(n)) &= \bigg(\bbP\big(\widehat{A}_{1,3} > \widehat{A}_{1,2}\big) + \bbP\big(\widehat{A}_{1,3} > \widehat{A}_{2,3}\big) \nonumber \\
&\qquad + \frac{1}{2}\,\bbP\big(\widehat{A}_{1,3} = \widehat{A}_{1,2}\big) + \frac{1}{2}\, \bbP\big(\widehat{A}_{1,3} = \widehat{A}_{2,3}\big) \bigg) \big(1+o(n^{-1})\big). \label{eq:ExAsy3node_v2}
\end{align}
Using Lem.~\ref{lem:ExactAsymp3nodes}, and the symmetry in $\cE_{P}$, we have
\begin{align}
\bbP\big(\widehat{A}_{1,3} > \widehat{A}_{1,2}\big) \,&=\, \big(f(n) - \tilde{f}(n)\big)\big(1 + o(n^{-1})\big) \,=\, \bbP\big(\widehat{A}_{1,3} > \widehat{A}_{2,3}\big), \label{eq:ExAsy3node_P4} \\
\bbP\big(\widehat{A}_{1,3} = \widehat{A}_{1,2}\big) \,&=\, \tilde{f}(n)\big(1+o(n^{-1})\big) \,=\, \bbP\big(\widehat{A}_{1,3} = \widehat{A}_{2,3}\big) ,\label{eq:ExAsy3node_P5}
\end{align}
and the proposition is proved by combining \eqref{eq:ExAsy3node_v2}, \eqref{eq:ExAsy3node_P4}, and \eqref{eq:ExAsy3node_P5}.
\qed

\section{Proof of Theorem~\ref{thm:ExactAsymptotics}} \label{app:ExactAsymptotics}
We know that the dominant error in learning a tree distribution $P \in \cD(\cT^p, \theta)$ occurs at $3$-node sub-trees of $T_P$ (see Appendix~\ref{app:ErrExpVal}), and the corresponding error event has the following form: $\big\{\widehat{A}_{i,k} \ge \widehat{A}_{i,j}\big\}$ or $\big\{\widehat{A}_{i,k} \ge \widehat{A}_{j,k}\big\}$ where $\left\{ \{i,j\}, \{j,k\} \right\} \subset \cE_{P}$. The exponent corresponding to these dominant error events is the smallest among the set of all error events, and is given by $K_P$ in~\eqref{eq:ErrExpVal}. The exact error asymptotics for such a $3$-node sub-tree is given by Prop.~\ref{prop:ExactAsymp3nodes}, and hence the exact asymptotics for general $P \in \cD(\cT^p, \theta)$ is given by
\begin{equation}
\bbP(\cA_P(n)) = \kappa_P \big(2 f(n) - \tilde{f}(n)\big)(1+o(n^{-1})), \label{eq:ExactAsymptotics_v2}
\end{equation}
where $\kappa_P$ denotes the number of distinct $3$-node sub-trees of $T_P$ for which the corresponding error exponent is $K_P$. Comparing \eqref{eq:ExactAsymptotics} and \eqref{eq:ExactAsymptotics_v2}, observe that it only remains to show that $\kappa_P = \sum_{i=1}^p d_i(d_i-1)/2$. To characterize $\kappa_P$, we count the number of distinct $3$-node sub-trees of $T_P$ having the following form: a sub-tree with vertex set $\{i,j,k\} \subset \cV$ satisfying $\left\{\{i,j\}, \{j,k\}\right\} \subset \cE_P$. Towards this, for $j \in \cV = \{1,\ldots,p\}$, we recall that $\nbd(j)=\{i \in \cV : \{i,j\} \in \cE_{P} \}$ denotes the neighborhood of $j$ and additionally  define 
\begin{align*}
\cS_j &:= \{\{i,j,k\}  : i\in \nbd(j),\, k\in \nbd(j),\, i\neq k \}.
\end{align*}
Now, if $\{i_1,i_2,i_3\} \subset \cV$ is a $3$-node sub-tree of $\cT_P$ with $\left\{\{i_1,i_2\},\{i_2,i_3\}\right\} \subset \cE_{P}$, then we have $\{i_1,i_2,i_3\} \in \cS_{i_2}$. Similarly, for $j \in \cV$, each element of $\cS_j$ contains the nodes of a $3$-node sub-tree contributing to dominant errors with exponent $K_P$. Therefore, we have
\begin{equation}
\kappa_P = \big| \medcup_{j \in \cV} \cS_j\,\big| \, . \label{eq:kappa_val1}
\end{equation} 
Note that $|\nbd(j)| = d_j$, where $d_j$ denotes the degree of node $j$, and so for $j \in \cV$, we have
\begin{equation}
|\cS_j| = \binom{d_j}{2} = \frac{d_j(d_j-1)}{2} . \label{eq:Tj_Size}
\end{equation} 
As the graph is a tree, it follows that $\cS_i \cap \cS_j = \emptyset$ for $i \neq j$, and hence from \eqref{eq:kappa_val1}, \eqref{eq:Tj_Size}, we get
\begin{equation}
\kappa_P = \sum_{j \in \cV} |\cS_j| = \sum_{j=1}^p \frac{d_j(d_j-1)}{2} =\dP , \label{eq:kappa_val2}
\end{equation}
and the proof is complete by substituting \eqref{eq:kappa_val2} in \eqref{eq:ExactAsymptotics_v2}.
\qed

\section{Proof of Proposition~\ref{prop:EstimateEdgeSet_NS}} \label{app:EstimateEdgeSet_NS}
According to Prop.~\ref{prop:AnEquivalence}, a graphical model belonging to $\cD(\cT^p,\theta)$ is uniquely characterized by its edge set. For $p=3$, we have $|\cD(\cT^3,\theta)| = 3$, and we denote the elements of $\cD(\cT^p,\theta)$ as $U$, $V$, and $W$, with corresponding edge sets $\cE_U = \big\{ \{1,2\}, \{2,3\} \big\}$,\, $\cE_V = \big\{ \{1,3\}, \{2,3\} \big\}$, and $\cE_W = \big\{ \{1,2\}, \{1,3\} \big\}$. Given $n$ noisy samples $\bry^n$, let $L_U(\bry^n)$, $L_V(\bry^n)$, and $L_W(\bry^n)$ denote the log-likelihood functions assuming that the underlying graphical model is $U$, $V$, and $W$, respectively. Then, we have
\begin{alignat*}{2}
L_U(\bry^n) &= \sum_{\bry \in \cY^3} \widehat{P}^{(q)}(\bry) \log U^{(q)}(\bry); & \quad U^{(q)}(\bry) &= \sum_{\brx \in \cX^3} q^{\delta_{\brx,\bry}} (1-q)^{p - \delta_{\brx,\bry}} U(\brx),\\
L_V(\bry^n) &= \sum_{\bry \in \cY^3} \widehat{P}^{(q)}(\bry) \log V^{(q)}(\bry); &\quad V^{(q)}(\bry) &= \sum_{\brx \in \cX^3} q^{\delta_{\brx,\bry}} (1-q)^{p - \delta_{\brx,\bry}} V(\brx),\\
L_W(\bry^n) &= \sum_{\bry \in \cY^3} \widehat{P}^{(q)}(\bry) \log W^{(q)}(\bry);& \quad W^{(q)}(\bry) &= \sum_{\brx \in \cX^3} q^{\delta_{\brx,\bry}} (1-q)^{p - \delta_{\brx,\bry}} W(\brx),
\end{alignat*}
where $\delta_{\brx,\bry}$ denotes the Hamming distance between $\brx$ and $\bry$. The ML algorithm will choose $U$ as the underlying graphical model if $L_U(\bry^n) > \max\big\{L_V(\bry^n), L_W(\bry^n) \big\}$. 

We first analyze the condition $L_U(\bry^n) > L_V(\bry^n)$. Towards this, we can readily verify that $U^{(q)}(0,0,0) = U^{(q)}(1,1,1) = V^{(q)}(1,1,1) = V^{(q)}(0,0,0)$ and $U^{(q)}(0,1,1) = U^{(q)}(1,0,0) = V^{(q)}(1,0,0) = V^{(q)}(0,1,1)$. Also, we have
\begin{align*}
U^{(q)}(0,0,1) &= \left( (1-q)^3 + q^3 \right)\frac{\theta(1-\theta)}{2} + q(1-q) \left(\frac{1 - \theta(1-\theta)}{2}\right) = U^{(q)}(1,1,0), \\*
U^{(q)}(0,1,0) &= \left( (1-q)^3 + q^3 \right)\frac{\theta^2}{2} + q(1-q) \left(\frac{1 - \theta^2}{2}\right) = U^{(q)}(1,0,1).
\end{align*}
Further, we can verify that $V^{(q)}(1,1,0) = V^{(q)}(0,0,1) = U^{(q)}(0,1,0)$ and $V^{(q)}(1,0,1) = V^{(q)}(0,1,0) = U^{(q)}(0,0,1)$. Combining the above relations, we observe that the condition $L_U(\bry^n) > L_V(\bry^n)$ is equivalent to the following
\begin{equation*}
\left(\widehat{P}^{(q)}(0,0,1) + \widehat{P}^{(q)}(1,1,0)\right) \log\frac{U^{(q)}(0,0,1)}{U^{(q)}(0,1,0)} > \left(\widehat{P}^{(q)}(0,1,0) + \widehat{P}^{(q)}(1,0,1)\right) \log\frac{U^{(q)}(0,0,1)}{U^{(q)}(0,1,0)}.
\end{equation*}
As $0 < \theta < 0.5$ and $0 \le q < 0.5$, we have $(1-q)^3 + q^3 > q(1-q)$ and $1-\theta > \theta$, and hence it follows that $U^{(q)}(0,0,1) > U^{(q)}(0,1,0)$. Thus, we have
\begin{align*}
L_U(\bry^n) > L_V(\bry^n) &\iff \widehat{P}^{(q)}(0,0,1) + \widehat{P}^{(q)}(1,1,0) > \widehat{P}^{(q)}(0,1,0) + \widehat{P}^{(q)}(1,0,1) \\
&\iff \widehat{P}^{(q)}_{1,2}(0,0) + \widehat{P}^{(q)}_{1,2}(1,1) > \widehat{P}^{(q)}_{1,3}(0,0) + \widehat{P}^{(q)}_{1,3}(1,1) \\
&\iff \widehat{A}^{(q)}_{1,2} > \widehat{A}^{(q)}_{1,3}.
\end{align*}

Similarly, it can be shown that $L_U(\bry^n) > L_W(\bry^n)$ if and only if $\widehat{A}^{(q)}_{2,3} > \widehat{A}^{(q)}_{1,3}$. Therefore, the ML algorithm chooses the edge set $\cE_U = \big\{ \{1,2\}, \{2,3\} \big\}$ when $\widehat{A}^{(q)}_{1,2} > \widehat{A}^{(q)}_{1,3}$ and $\widehat{A}^{(q)}_{2,3} > \widehat{A}^{(q)}_{1,3}$. It follows from symmetry that the ML algorithm chooses the edge set $\big\{ \{i,j\}, \{j,k\} \big\}$ when $\widehat{A}^{(q)}_{i,j} > \widehat{A}^{(q)}_{i,k}$ and $\widehat{A}^{(q)}_{j,k} > \widehat{A}^{(q)}_{i,k}$. This is equivalent to the ML algorithm choosing the edge set of an MWST over a complete weighted graph where the weight of the edge between nodes $i$ and $j$ is equal to $\widehat{A}^{(q)}_{i,j}$.
\qed

\section{Proof of Theorem~\ref{thm:KPq_Val}} \label{app:KPq_Val}
Towards proving the theorem, we first prove the following lemma for $p=3$.
\begin{lemma} \label{lem:KPq_Val}
	When $p=3$, $P \in \cD(\cT^3,\theta)$, and $P^{(q)}$ is given by~\eqref{eq:Dist_Pq}, then the error exponent $K_P^{(q)}$ is given by \eqref{eq:KPq_Val}.
\end{lemma}
\begin{IEEEproof}
	Without loss of generality, assume that $\cE_P = \big\{ \{1,2\}, \{2,3\} \big\}$. When $n$ i.i.d. samples are drawn from $P^{(q)}$, the error event $\big\{\cE^{(q)}(\bry^n) \neq \cE_P \big\}$ occurs when either $\widehat{A}_{1,3}^{(q)} \ge \widehat{A}_{1,2}^{(q)}$ or $\widehat{A}_{1,3}^{(q)} \ge \widehat{A}_{2,3}^{(q)}$. From symmetry, we have $\bbP\big(\widehat{A}_{1,3}^{(q)} \ge \widehat{A}_{1,2}^{(q)}\big) =  \bbP\big(\widehat{A}_{1,3}^{(q)} \ge \widehat{A}_{1,2}^{(q)}\big)$, and hence
	\begin{equation}
	K_{P}^{(q)} = \lim_{n \to \infty} -\frac{1}{n} \log \bbP\big(\widehat{A}_{1,3}^{(q)} \ge \widehat{A}_{1,2}^{(q)}\big) .\label{eq:KPq_Val_v2}
	\end{equation} 
	As $\widehat{A}_{i,j}^{(q)} = \widehat{P}_{i,j}^{(q)}(0,0) + \widehat{P}_{i,j}^{(q)}(1,1)$, it follows from Sanov's theorem~\cite[Thm.~11.4.1]{CoverBook06} that
	\begin{equation}
	K_{P}^{(q)} = \min_{Q \in \cP(\cY^3)} \left\{D(Q \| P^{(q)}) : Q_{1,3}(0,0) + Q_{1,3}(1,1) \ge Q_{1,2}(0,0) + Q_{1,2}(1,1) \right\}. \label{eq:KPq_Val_v3}
	\end{equation}
	The constraint $Q_{1,3}(0,0) + Q_{1,3}(1,1) \ge Q_{1,2}(0,0) + Q_{1,2}(1,1)$ is equivalent to the following:
	\begin{equation}
	Q(0,1,0)+Q(1,0,1) \ge Q(0,0,1)+Q(1,1,0). \label{eq:SanovConstraint_v2}
	\end{equation}
	Let $Q^*$ denote that $Q \in \cP(\cY^3)$ which satisfies \eqref{eq:SanovConstraint_v2} and minimizes $D(Q\| P^{(q)})$, and let $g$ be the map $g : \cY^3 \to \bbR$ given by~\eqref{eq:Define_g}.
	Then, using a Lagrange multiplier, $Q^*$ can be obtained as the exponentially tilted distribution
	\begin{equation}
	Q^*(\bry) = \frac{P^{(q)}(\bry) \exp\left(\lambda g(\bry)\right)}{\sum_{\tilde{\bry} \in \cY^3} P^{(q)}(\tilde{\bry}) \exp\left(\lambda g(\tilde{\bry})\right)}, ~~\bry \in \cY^3, \label{eq:TiltedDistQ_v2}
	\end{equation}
	where $\lambda$ is chosen to satisfy \eqref{eq:SanovConstraint_v2}. Using~\eqref{eq:Dist_Pq}, it can be verified that $P^{(q)}(0,0,1) = \beta_1$ (see~\eqref{eq:beta1}) and $P^{(q)}(0,1,0) = \beta_2$ (see~\eqref{eq:beta2}), and it follows that 
	\begin{equation}
	\lambda = \frac{1}{2} \log\left(\frac{P^{(q)}(0,0,1)}{P^{(q)}(0,1,0)}\right) = \frac{1}{2} \log\left(\frac{\beta_1}{\beta_2}\right). \label{eq:lambdaVal_v2}
	\end{equation}
	Now, using the fact that $K_P^{(q)} = D(Q^* \| P^{(q)})$, and substituting~\eqref{eq:lambdaVal_v2} in~\eqref{eq:TiltedDistQ_v2}, we obtain 
	\begin{equation}
	K_P^{(q)} = -\log\left(1 - 4\Big(\frac{\beta_1 + \beta_2}{2} - \sqrt{\beta_1 \beta_2} \Big)\right)  \label{eq:KPq_Val_v4}
	\end{equation}	
	as desired.
\end{IEEEproof}
The above lemma shows that Thm.~\ref{thm:KPq_Val} holds when  $P$ has $3$ nodes. For $p>3$ nodes, it is known that the dominant error event in the learning problem using noisy samples occurs at various $3$-node sub-trees corresponding to nodes $\{i,j,k\} \subset \cV$ satisfying $\left\{\{i,j\}, \{j,k\}\right\} \subset \cE_P$ (see \cite[Sec.~4, App.~A.1]{Nikolakakis19NonParametric}). This observation is related to the correlation decay property for tree models with uniform marginals over a binary alphabet~\cite[Lem.~A.2]{Nikolakakis19Predictive}. For a general $p$-node tree, the number of such sub-trees  resulting in dominant error  is equal to $\dP$ in~\eqref{eq:dP_Def}, and Lem.~\ref{lem:KPq_Val} shows that the exponent corresponding to this dominant error term is given by \eqref{eq:KPq_Val_v4}. As the error exponent only depends on the dominant error term, it follows that the error exponent using noisy samples, when $P \in \cD(\cT^p,\theta)$ and $p>3$, is also given by \eqref{eq:KPq_Val_v4}. 
\qed

\section{Proof of Theorem~\ref{thm:ExactAsymp_NS}} \label{app:ExactAsymp_NS}
The following lemma states the result for the special case of $p=3$ nodes.
\begin{lemma} \label{lem:ExactAsymp3nodes_NS}
	When $P \in \cD(\cT^3, \theta)$, and $\bry^n$ are $n$ i.i.d. samples distributed according to $P^{(q)}$ in~\eqref{eq:Dist_Pq}, then we have
	\begin{equation}
	\bbP\big(\hat{\cE}^{(q)}(\bry^n) \neq \cE_{P}\big) \,=\, \big(2 f^{(q)}(n) - \tilde{f}^{(q)}(n)\big)\big(1 + o(n^{-1})\big), \label{eq:ExactAsymp3nodes_NS}
	\end{equation}
	where $\tilde{f}^{(q)}(n)$ and $f^{(q)}(n)$ are given by~\eqref{eq:tilde_fq_Def} and~\eqref{eq:fq_Def}, respectively. 
\end{lemma}
\begin{IEEEproof}
	For a $3$-node tree, we assume without loss of generality that $\cE_{P} = \big\{\{1,2\},\{2,3\} \big\}$. When an implementation of the MWST algorithm (such as Prim's algorithm~\cite{AlgoBook09}) randomly breaks ties during the construction of the MWST, then similar to the noiseless samples setting analyzed in~\eqref{eq:ExAsy3node_v1}, we have that
	\begin{align}
	\bbP\big(\hat{\cE}^{(q)}(\bry^n) \neq \cE_{P}\big) &= \bbP\big(\big\{\widehat{A}_{1,3}^{(q)} > \widehat{A}_{1,2}^{(q)}\big\}  \cup \big\{\widehat{A}_{1,3}^{(q)} > \widehat{A}_{2,3}^{(q)}\big\} \big) + \frac{1}{2}\, \bbP\big(\widehat{A}_{2,3}^{(q)} > \widehat{A}_{1,2}^{(q)} = \widehat{A}_{1,3}^{(q)} \big) \nonumber \\
	&\qquad + \frac{1}{2}\, \bbP\big(\widehat{A}_{1,2}^{(q)} > \widehat{A}_{2,3}^{(q)} = \widehat{A}_{1,3}^{(q)} \big) + \frac{2}{3}\, \bbP\big(\widehat{A}_{1,2}^{(q)} = \widehat{A}_{2,3}^{(q)} = \widehat{A}_{1,3}^{(q)} \big). \label{eq:ExAsy3node_NS_v1}
	\end{align}
	Now, collecting the error events corresponding to the smallest exponent, we obtain
	\begin{align}
	\bbP\big(\hat{\cE}^{(q)}(\bry^n) \neq \cE_{P}\big) \,&=\, \bigg(\bbP\big(\widehat{A}_{1,3}^{(q)} > \widehat{A}_{1,2}^{(q)}\big) + \bbP\big(\widehat{A}_{1,3}^{(q)} > \widehat{A}_{2,3}^{(q)}\big) \nonumber \\*
	&\qquad + \frac{1}{2}\,\bbP\big(\widehat{A}_{1,3}^{(q)} = \widehat{A}_{1,2}^{(q)}\big) + \frac{1}{2}\, \bbP\big(\widehat{A}_{1,3}^{(q)} = \widehat{A}_{2,3}^{(q)}\big) \bigg) \big(1+o(n^{-1})\big). \label{eq:ExAsy3node_NS_v2}
	\end{align}
	By symmetry, we have $\bbP\big(\widehat{A}_{1,3}^{(q)} > \widehat{A}_{2,3}^{(q)}\big) = \bbP\big(\widehat{A}_{1,3}^{(q)} > \widehat{A}_{1,2}^{(q)}\big)$ and $\bbP\big(\widehat{A}_{1,3}^{(q)} = \widehat{A}_{2,3}^{(q)}\big) = \bbP\big(\widehat{A}_{1,3}^{(q)} = \widehat{A}_{1,2}^{(q)}\big)$, and hence from \eqref{eq:ExAsy3node_NS_v2}, we get
	\begin{equation}
	\bbP\big(\hat{\cE}^{(q)}(\bry^n) \neq \cE_{P}\big) \,=\, \Big( 2\,\bbP\big(\widehat{A}_{1,3}^{(q)} \ge \widehat{A}_{1,2}^{(q)}\big) \,-\, \bbP\big(\widehat{A}_{1,3}^{(q)} = \widehat{A}_{2,3}^{(q)}\big) \Big) \big(1+o(n^{-1})\big). \label{eq:ExAsy3node_NS_v3}
	\end{equation}
	Note that the observed sample $\bry$ has distribution $P^{(q)}$, and takes values in alphabet $\cY^3 = \{0,1\}^3$. Now, if we define the random variable $U_1$ as
	\begin{equation*}
	U_1 = \begin{cases}
	1, &\mathrm{~if~} \bry \in \left\{(0,1,0), (1,0,1) \right\}, \\
	-1 &\mathrm{~if~} \bry \in \left\{(0,0,1), (1,1,0) \right\}, \\
	0 &\mathrm{~if~} \bry \in \left\{(0,0,0), (1,1,1), (0,1,1), (1,0,0) \right\},
	\end{cases}
	\end{equation*}
	and consider $n$ i.i.d. variables $\{U_i\}_{i=1}^n$, then we observe that 
	\begin{align}
	\left\{ \widehat{A}_{1,3}^{(q)} \ge \widehat{A}_{1,2}^{(q)} \right\} \,&=\, \left\{\widehat{P}^{(q)}(0,1,0) + \widehat{P}^{(q)}(1,0,1) - \widehat{P}^{(q)}(0,0,1) - \widehat{P}^{(q)}(1,1,0) \ge 0 \right\} \nonumber\\*
	&=\, \bigg\{\sum_{i=1}^n U_i \ge 0 \bigg\}. \label{eq:EquivalenceOfErrorEvents}
	\end{align}  
	It can be verified that $\widehat{P}^{(q)}(0,0,1) = \widehat{P}^{(q)}(1,1,0) = \beta_1$, given by~\eqref{eq:beta1}, while $\widehat{P}^{(q)}(0,1,0) = \widehat{P}^{(q)}(1,0,1) = \beta_2$, given by~\eqref{eq:beta2}. As $\beta_1 > \beta_2$, it follows that $\bbE[U_1] < 0$. Further, we have $\bbP\big(\sum_{i=1}^n U_i=0\big) > 0$ for every admissible $n$, and hence the conditions for applying the strong large deviations results in~\cite{BlackwellHodges59} are satisfied. Note that $U_1$ takes values in the alphabet $\cU =\{-1,0,1\}$, and we denote the probability distribution of $U_1$ as $Q(u)$, $u \in \cU$. Consider the moment generating function of $U_1$ defined as
	\begin{equation*}
	\phi(t) := \bbE_Q\left[e^{t U_1}\right] = Q(-1)\,e^{-t} + Q(0) + Q(1)\,e^t,
	\end{equation*}
	and let $\tau$ be the value of $t$ which minimizes $\phi(t)$, i.e. $\tau = \argmin_{t\in\mathbb{R}} \phi(t)$. Then, $\tau$ is uniquely determined as the solution of $\phi^{\prime}(\tau) = 0$, which gives us 
	\begin{align}
	e^{\tau} &= \sqrt{\frac{Q(-1)}{Q(1)}} = \sqrt{\frac{\beta_1}{\beta_2}}, \label{eq:tau_val_NS}\\
	\phi(\tau) &= Q(0) + \sqrt{4 Q(-1) Q(1)} = \big(1-(2\beta_1 + 2\beta_2)\big) + 4\sqrt{\beta_1 \beta_2}= \exp\big(-K_P^{(q)}\big), \label{eq:PhiTau_and_KPq}
	\end{align}
	where \eqref{eq:PhiTau_and_KPq} follows from \eqref{eq:KPq_Val}. Now, define a random variable $V_1$ taking values in $\cU$ having a tilted distribution, denoted $\tilde{Q}$, and defined as $\tilde{Q}(u) := Q(u)\, e^{\tau u}/\phi(\tau)$, $u \in \cU$.
	Note that $\bbE_{\tilde{Q}}[V_1] = \phi^{\prime}(\tau)/\phi(\tau) = 0$. Let $\mu_2$, $\mu_3$, and $\mu_4$ denote the second, third, and fourth central moments of $V_1$, respectively. Then, we have
	\begin{align*}
	\mu_2 &= \frac{\phi^{\prime \prime}(\tau)}{\phi(\tau)} = \frac{Q(-1)e^{-\tau} + Q(1) e^{\tau}}{\phi(\tau)} = 4 \sqrt{\beta_1 \beta_2}\, \exp(K_P^{(q)}) ,\\
	\mu_3 &= \frac{\phi^{\prime\prime \prime}(\tau)}{\phi(\tau)} = \frac{\phi^{\prime}(\tau)}{\phi(\tau)} = 0 , \qquad \quad \mu_4 = \frac{\phi^{\prime\prime \prime\prime}(\tau)}{\phi(\tau)} = \frac{\phi^{\prime\prime}(\tau)}{\phi(\tau)} = \mu_2 .
	\end{align*}
	From the strong large deviations theorem~\cite[Thm.~3]{BlackwellHodges59}, we have
	\begin{equation}
	\bbP \bigg(\sum_{i=1}^n U_i = 0\bigg) \, = \, \frac{[\phi(\tau)]^n}{\sqrt{2 \pi \mu_2 n}} \left[1 + \frac{1}{8n}\left(\frac{\mu_4}{\mu_2^2} -3 -\frac{5 \mu_3^2}{3 \mu_2^3} \right) \right]\left( 1 + o(n^{-1})\right). \label{eq:ExactAsymp_NS_eq_BH}
	\end{equation}
	Using the fact that $\bbP\big(\sum_{i=1}^n U_i = 0\big) \,=\, \bbP\big(\widehat{A}_{1,3}^{(q)} = \widehat{A}_{1,2}^{(q)}\big)$, and substituting the values of $\phi(\tau)$, $\mu_2$, $\mu_3$, and $\mu_4$, in \eqref{eq:ExactAsymp_NS_eq_BH}, we obtain
	\begin{equation}
	 \bbP\big(\widehat{A}_{1,3}^{(q)} = \widehat{A}_{1,2}^{(q)}\big) \,=\, \tilde{f}^{(q)}(n) \left(1 + o(n^{-1}) \right) , \label{eq:ExactAsymp3nodes_NS_eq_v2}
	\end{equation}
	where $\tilde{f}^{(q)}(n)$ is given by \eqref{eq:tilde_fq_Def}. From the strong large deviations theorem, as stated in~\cite[Thm.~4]{BlackwellHodges59}, we have
	\begin{equation}
	\bbP \bigg(\sum_{i=1}^n U_i \ge 0\bigg) \,=\, \frac{\tilde{f}^{(q)}(n)}{1-z} \left[1 - \frac{1}{2n}\left(\frac{(z \mu_3/\mu_2) + z(1+z)/(1-z)}{(1-z) \mu_2}\right)\right] \left(1 + o(n^{-1})\right) , \label{eq:ExactAsymp3nodes_NS_ge_v2}
	\end{equation}
	where $z = e^{-\tau} = \sqrt{\beta_2/\beta_1}$. Now, combining \eqref{eq:EquivalenceOfErrorEvents} and \eqref{eq:ExactAsymp3nodes_NS_ge_v2}, and the fact that $\mu_3 = 0$, we obtain
	\begin{equation}
	\bbP\big(\widehat{A}_{1,3}^{(q)} \ge \widehat{A}_{1,2}^{(q)}\big) \,=\, f^{(q)}(n) \left(1 + o(n^{-1})\right) , \label{eq:ExactAsymp3nodes_NS_ge_v3}
	\end{equation}
	where $f^{(q)}(n)$ is given by \eqref{eq:fq_Def}. Finally, we obtain \eqref{eq:ExactAsymp3nodes_NS}  by combining \eqref{eq:ExAsy3node_NS_v3}, \eqref{eq:ExactAsymp3nodes_NS_eq_v2}, and \eqref{eq:ExactAsymp3nodes_NS_ge_v3}.
\end{IEEEproof}

The above lemma provides exact asymptotics using noisy samples, when the underlying graphical is over $3$-nodes. For $p>3$ nodes, it is known that the dominant error event in the learning problem using noisy samples occurs at various $3$-node sub-trees corresponding to nodes $\{i,j,k\} \subset \cV$ satisfying $\left\{\{i,j\}, \{j,k\}\right\} \subset \cE_P$ (see \cite[Sec.~4, App.~A.1]{Nikolakakis19NonParametric}). The exponent corresponding to these dominant error events is the smallest among the set of all error events, and is given by $K_P^{(q)}$ in~\eqref{eq:KPq_Val}. The exact error asymptotics for such a $3$-node sub-tree is given by~\eqref{eq:ExactAsymp3nodes_NS}, and hence the exact asymptotics for a general $p$-node tree, with underlying graphical model $P \in \cD(\cT^p, \theta)$, is given by
\begin{equation*}
\bbP\big(\hat{\cE}^{(q)}(\bry^n) \neq \cE_{P}\big) \,=\, \dP\,\big(2 f^{(q)}(n) - \tilde{f}^{(q)}(n)\big)\big(1 + o(n^{-1})\big),
\end{equation*}
where $\dP$ in~\eqref{eq:dP_Def} is the number of such $3$-node sub-trees contributing to dominant error events, with corresponding exponent $K_P^{(q)}$.
\qed

\section{Extremal tree structures: Star and Markov chain} \label{app:StarMC}
We show for a $p$-node tree-structured graphical model $P \in \cD(\cT^p,\theta)$,  the star and the Markov chain tree structures are extremal in the following sense.
\begin{proposition} \label{prop:extremal}
	For $P \in \cD(\cT^p,\theta)$ with $p>3$, the value of $\dP$  in~\eqref{eq:dP_Def} is maximized (resp.\ minimized) when the underlying tree structure is a star (resp.\ Markov chain).
\end{proposition}
\begin{IEEEproof}
For a given $P \in \cD(\cT^p,\theta)$, let $i_1, \ldots, i_p$ be a permutation of $1,\ldots,p$ such that $d_{i_1} \ge d_{i_2} \ge \cdots \ge d_{i_p}$, where $d_j$ denotes the degree of node $j$. Note that star and Markov chain tree structures are characterized (up to isomorphism) by the following property: $d_{i_1} = p-1$ for a star, while $d_{i_1}=2$ for a Markov chain. For proving their extremal nature, we will use the following inequality, where for any given real numbers $v_1 \ge v_2 > 1$,
\begin{equation}
v_1 (v_1 - 1) + v_2 (v_2 - 1) < (v_1+1) v_1 + (v_2-1)(v_2-2). \label{eq:StarMC}
\end{equation}  

--  If the underlying tree is not a star, then there exists a node $j \neq i_1$ with $1 < d_j < p-1$. If $\{j,k\} \in \cE_{P}$. Using~\eqref{eq:dP_Def} and~\eqref{eq:StarMC}, we observe that when edge $\{j,k\}$ is replaced by the edge $\{i_1,k\}$, then $\dP$ is increased for the modified tree structure. This process of increasing $\dP$ can be repeated until the resulting tree structure is a star, i.e., $d_{i_1} = p-1$ and $d_{i_2}=\ldots=d_{i_p}=1$.

-- On the other hand, if the underlying tree is not a Markov chain, then $d_{i_1} > 2$. Let $\{i_1,k\} \in \cE_{P}$ with $k \neq i_p$. Then, by replacing  the edge $\{i_1,k\}$ with the edge  $\{k,i_p\}$, and using~\eqref{eq:dP_Def} and~\eqref{eq:StarMC}, we observe that the value of $\dP$ is decreased for the modified tree structure. This process of decreasing $\dP$ can be repeated until the resulting tree structure is a Markov chain.
 \end{IEEEproof}
Note that the star and the Markov chains structures coincide for a $3$-node tree. We additionally note that Prop.~\ref{prop:extremal}, together with Thm.~\ref{thm:ExactAsymptotics} and Thm.~\ref{thm:ExactAsymp_NS}, imply that for a fixed $\theta$, the star and Markov chain structures  are extremal in terms of the  error probabilities in learning them. This observation  holds for all $0<\theta<0.5$ whereas the corresponding result concerning extremal structures in~\cite{Tan10TSP} holds for a rather restrictive set of  correlation parameters.

 


\begin{thebibliography}{10}
	\providecommand{\url}[1]{#1}
	\csname url@samestyle\endcsname
	\providecommand{\newblock}{\relax}
	\providecommand{\bibinfo}[2]{#2}
	\providecommand{\BIBentrySTDinterwordspacing}{\spaceskip=0pt\relax}
	\providecommand{\BIBentryALTinterwordstretchfactor}{4}
	\providecommand{\BIBentryALTinterwordspacing}{\spaceskip=\fontdimen2\font plus
		\BIBentryALTinterwordstretchfactor\fontdimen3\font minus
		\fontdimen4\font\relax}
	\providecommand{\BIBforeignlanguage}[2]{{%
			\expandafter\ifx\csname l@#1\endcsname\relax
			\typeout{** WARNING: IEEEtran.bst: No hyphenation pattern has been}%
			\typeout{** loaded for the language `#1'. Using the pattern for}%
			\typeout{** the default language instead.}%
			\else
			\language=\csname l@#1\endcsname
			\fi
			#2}}
	\providecommand{\BIBdecl}{\relax}
	\BIBdecl
	
	\bibitem{Wainwright08_FnT}
	M.~J. Wainwright and M.~I. Jordan, ``Graphical models, exponential families,
	and variational inference,'' \emph{Found. Trends Mach. Learn.}, vol.~1, no.
	1-2, pp. 1--305, 2008.
	
	\bibitem{Besag86}
	J.~Besag, ``On the statistical analysis of dirty pictures,'' \emph{J. Roy.
		Statist. Soc., Ser. B}, vol.~48, no.~3, pp. 259--302, 1986.
	
	\bibitem{Kschischang98}
	F.~R. {Kschischang} and B.~J. {Frey}, ``Iterative decoding of compound codes by
	probability propagation in graphical models,'' \emph{{IEEE} J. Sel. Areas
		Commun.}, vol.~16, no.~2, pp. 219--230, 1998.
	
	\bibitem{Navigli10_PAMI}
	R.~{Navigli} and M.~{Lapata}, ``An experimental study of graph connectivity for
	unsupervised word sense disambiguation,'' \emph{{IEEE} Trans. Pattern Anal.
		Mach. Intell.}, vol.~32, no.~4, pp. 678--692, 2010.
	
	\bibitem{Maneva07}
	E.~Maneva, E.~Mossel, and M.~J. Wainwright, ``A new look at survey propagation
	and its generalizations,'' \emph{Journal of the {ACM}}, vol.~54, no.~4, pp.
	2--41, 2007.
	
	\bibitem{ChowLiu68}
	C.~K. {Chow} and C.~N. {Liu}, ``Approximating discrete probability
	distributions with dependence trees,'' \emph{{IEEE} Trans. Inform. Theory},
	vol.~14, no.~3, pp. 462--467, May 1968.
	
	\bibitem{Csiszar03_IP}
	I.~Csisz\'{a}r and F.~Mat\'{u}\u{s}, ``Information projections revisited,''
	\emph{{IEEE} Trans. Inform. Theory}, vol.~49, no.~6, pp. 1474--1490, Jun.
	2003.
	
	\bibitem{ChowWagner73}
	C.~K. {Chow} and T.~J. {Wagner}, ``Consistency of an estimate of tree-dependent
	probability distributions,'' \emph{{IEEE} Trans. Inform. Theory}, vol.~19,
	no.~3, pp. 369--371, May 1973.
	
	\bibitem{HollanderBook00}
	F.~den Hollander, \emph{Large Deviations}, ser. Fields Institute
	Monographs.\hskip 1em plus 0.5em minus 0.4em\relax American Mathematical
	Soc., 2000.
	
	\bibitem{CoverBook06}
	{T.~M.~Cover and J.~A.~Thomas}, \emph{Elements of Information Theory},
	2nd~ed.\hskip 1em plus 0.5em minus 0.4em\relax Hoboken, N.J.:
	Wiley-Interscience, 2006.
	
	\bibitem{Tan11IT}
	V.~Y.~F. {Tan}, A.~{Anandkumar}, L.~{Tong}, and A.~S. {Willsky}, ``A
	large-deviation analysis of the maximum-likelihood learning of {Markov} tree
	structures,'' \emph{{IEEE} Trans. Inform. Theory}, vol.~57, no.~3, pp.
	1714--1735, Mar. 2011.
	
	\bibitem{Tan10TSP}
	V.~Y.~F. {Tan}, A.~{Anandkumar}, and A.~S. {Willsky}, ``Learning {Gaussian}
	tree models: {Analysis} of error exponents and extremal structures,''
	\emph{{IEEE} Trans. Signal Process.}, vol.~58, no.~5, pp. 2701--2714, May
	2010.
	
	\bibitem{BahadurRao60}
	R.~R. Bahadur and R.~Ranga~Rao, ``On deviations of the sample mean,''
	\emph{Ann. Math. Statist.}, vol.~31, no.~4, pp. 1015--1027, Dec. 1960.
	
	\bibitem{BlackwellHodges59}
	D.~Blackwell and J.~L. Hodges, ``The probability in the extreme tail of a
	convolution,'' \emph{Ann. Math. Statist.}, vol.~30, no.~4, pp. 1113--1120,
	Dec. 1959.
	
	\bibitem{BreslerKarzand18}
	G.~Bresler and M.~Karzand, ``Learning a tree-structured {Ising} model in order
	to make predictions,'' \emph{Ann. Statist.}, 2020, arXiv:1604.06749v3
	[cs.ST].
	
	\bibitem{Haim18}
	E.~Haim, Y.~Kochman, and U.~Erez, ``On random-coding union bounds with and
	without erasures,'' \emph{{IEEE} Trans. Inform. Theory}, vol.~64, no.~6, pp.
	4294--4308, Jun 2018.
	
	\bibitem{Nikolakakis19AISTATS}
	K.~E. Nikolakakis, D.~S. Kalogerias, and A.~D. Sarwate, ``Learning tree
	structures from noisy data,'' in \emph{Proc. AISTATS}, Naha, Okinawa, Japan,
	2019, pp. 1771--1782.
	
	\bibitem{Nikolakakis19Predictive}
	------, ``Predictive learning on hidden tree-structured {Ising} models,'' Feb.
	2019, arXiv:1812.04700v2 [stat.ML].
	
	\bibitem{Nikolakakis19NonParametric}
	------, ``Non-parametric structure learning on hidden tree-shaped
	distributions,'' Sep. 2019, arXiv:1909.09596v1 [stat.ML].
	
	\bibitem{Cheng18NeurIPS}
	Y.~Cheng, I.~Diakonikolas, D.~M. Kane, and A.~Stewart, ``Robust learning of
	fixed-structure {Bayesian} networks,'' in \emph{Proc. NeurIPS}, Montreal,
	Canada, 2018, pp. 10\,304--10\,316.
	
	\bibitem{LauritzenBook}
	S.~Lauritzen, \emph{Graphical Models}.\hskip 1em plus 0.5em minus 0.4em\relax
	{Oxford, U.K.}: {Oxford Univ. Press}, 1996.
	
	\bibitem{HersteinBook75}
	I.~Herstein, \emph{{Topics In Algebra}}, 2nd~ed.\hskip 1em plus 0.5em minus
	0.4em\relax John Wiley and Sons, New York, 1975.
	
	\bibitem{Shanmugam14NeurIPS}
	R.~Tandon, K.~Shanmugam, P.~Ravikumar, and A.~G. Dimakis, ``On the information
	theoretic limits of learning {Ising} models,'' in \emph{Proc. NeurIPS},
	Montreal, Canada, 2014, pp. 2303--2311.
	
	\bibitem{Anandkumar12}
	A.~Anandkumar, V.~Y.~F. Tan, F.~Huang, and A.~S. Willsky, ``High-dimensional
	structure estimation in {I}sing models: {L}ocal separation criterion,''
	\emph{Ann. Statist.}, vol.~40, no.~3, pp. 1346--1375, 2012.
	
	\bibitem{AlgoBook09}
	T.~H. Cormen, C.~E. Leiserson, R.~L. Rivest, and C.~Stein, \emph{Introduction
		to Algorithms}, 3rd~ed.\hskip 1em plus 0.5em minus 0.4em\relax {Cambridge,
		M.A.}: {The MIT Press}, 2009.
	
	\bibitem{Kester86}
	A.~D.~M. Kester and W.~C.~M. Kallenberg, ``Large deviations of estimators,''
	\emph{Ann. Statist.}, vol.~14, no.~2, pp. 648--664, 1986.
	
	\bibitem{Choi11}
	M.~J. Choi, V.~Y.~F. Tan, A.~Anandkumar, and A.~S. Willsky, ``Learning latent
	tree graphical models,'' \emph{J. Mach. Learn. Res.}, vol.~12, pp.
	1771--1812, 2011.
	
	\bibitem{Bresler15STOC}
	G.~Bresler, ``Efficiently learning {I}sing models on arbitrary graphs,'' in
	\emph{Proc. ACM Symp. Theory Comp. (STOC)}, New York, NY, USA, 2015, pp.
	771--782.
	
	\bibitem{Dasarathy16AISTATS}
	G.~Dasarathy, A.~Singh, M.-F. Balcan, and J.~H. Park, ``Active learning
	algorithms for graphical model selection,'' in \emph{Proc. AISTATS}, Cadiz,
	Spain, 2016, pp. 1356--1364.
	
	\bibitem{Scarlett17AISTATS}
	J.~Scarlett and V.~Cevher, ``Lower bounds on active learning for graphical
	model selection,'' in \emph{Proc. AISTATS}, Fort Lauderdale, Flordia, USA,
	2017, pp. 1356--1364.
	
	\bibitem{Moulin17}
	P.~{Moulin}, ``The log-volume of optimal codes for memoryless channels,
	asymptotically within a few nats,'' \emph{{IEEE} Trans. Inform. Theory},
	vol.~63, no.~4, pp. 2278--2313, Apr. 2017.
	
	\bibitem{Topsoe04}
	F.~Tops{\o}e, ``Some bounds for the logarithmic function,'' \emph{RGMIA Res.
		Rep. Collection}, vol.~7, no.~2, 2004.
	
\end{thebibliography}
\end{document}